\documentclass{article} 
\usepackage{iclr2027_conference,times}

\usepackage{amsmath,amsfonts,bm}

\def\eqref#1{equation~\ref{#1}}

\def\1{\bm{1}}

\DeclareMathAlphabet{\mathsfit}{\encodingdefault}{\sfdefault}{m}{sl}
\SetMathAlphabet{\mathsfit}{bold}{\encodingdefault}{\sfdefault}{bx}{n}

\usepackage{hyperref}
\usepackage{url}
\usepackage{booktabs}
\usepackage{amsmath}
\usepackage{amssymb}
\usepackage{graphicx}
\usepackage{xcolor}

\title{RESYNC: Residual Latent-State Synchronization for Deadline-Bound Streaming Inference}

\author{Eljas Linna, Hao Dong, Juho Kanniainen}

\iclrfinalcopy 
\begin{document}

\maketitle

\begin{abstract}
Predicting every observation in a high-rate stream requires a model to finish before the next arrival. A large model may offer better predictions but miss that deadline. We propose \textbf{RESYNC}: the large model runs asynchronously, while a fast path updates its latest exact but stale latent state using the observations received since that state was computed and applies its unchanged prediction head. The updater learns from the frozen model's representations without task labels. Across one real and two synthetic streams and five architecture families, RESYNC beats the best compact or distilled alternative with a paired 95\% interval above zero in all twelve dataset--architecture pairs with a material teacher advantage at anchor ages of 4--12 observations, and in eleven pairs at older tested ages up to 32. The share of the teacher's gain preserved remains complete on the real stream but is lower and generally declines with anchor age on the synthetic streams. In live replay with 1 ms arrivals, the synchronous large models cannot deliver any predictions on time while RESYNC serves at least 99.4\% of arrivals within the interval, with 0.35--0.62 ms median latency on one CPU thread: around the same speed as or faster than the compact model, and 7.9--17.7$\times$ faster than synchronous large-model inference. RESYNC targets settings with background compute available but a strict per-observation response deadline.
\end{abstract}

\section{Introduction}

Many streaming applications require a prediction for every incoming observation within a short deadline $D$. When observations arrive every $\Delta$ and a model needs longer than $\Delta$ to serve one, arrivals outpace completions: the backlog grows, and each prediction arrives only after later observations. We therefore study \emph{per-observation reaction latency in high-rate data streams}, the time from an observation's arrival to its prediction, together with coverage and deadline misses. Latency-aware evaluations of streaming perception have exposed the related mismatch between offline accuracy and the quality of the prediction actually available at a given time~\citep{li-2020-towards,chu-2026-bridging}. The same issue arises in window-based prediction on financial, physiological, industrial and other multivariate streams.

The usual response is to make the complete predictor fast enough, by reducing its capacity or context, compressing or distilling it, or exiting early~\citep{teerapittayanon-2016-branchynet,huang-2018-multiscale}. We call the high-capacity predictor the \emph{teacher}, and a smaller-capacity variant of the same architecture, run on every current window, the \emph{compact model}. A compact model can meet a short deadline but may sacrifice quality. Meanwhile, the teacher may already have computed an exact representation of a recent window. This representation becomes stale as observations arrive, but it still encodes long-context analysis that the compact model cannot recover before the deadline.

\begin{figure}[t]
\centering
\includegraphics[width=0.92\textwidth]{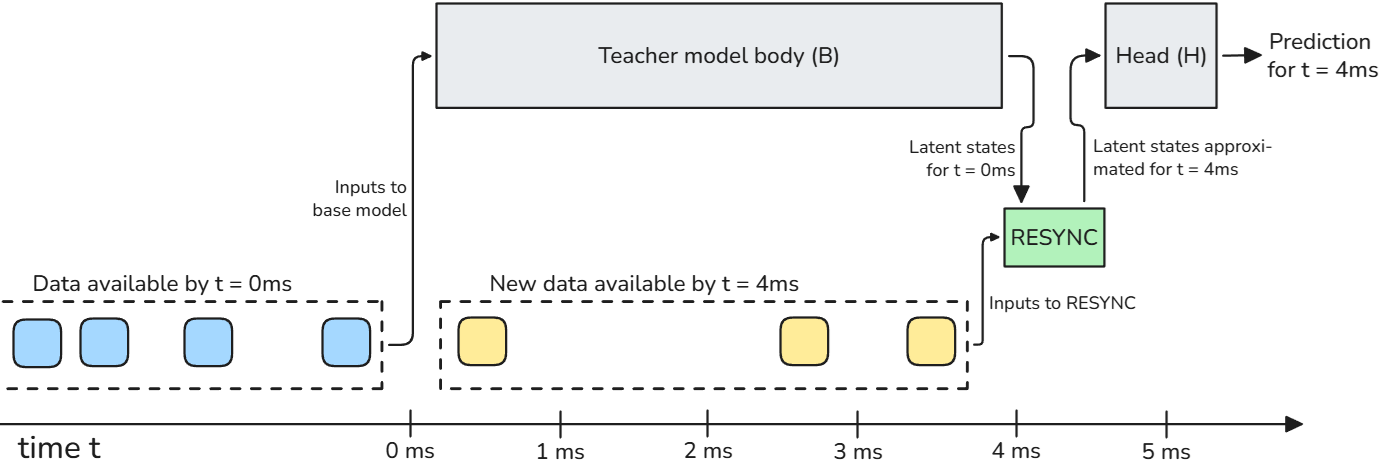}
\caption{RESYNC at one arrival. The teacher's body (B) processed the window available at $t=0$ ms and delivers its latent at $t=4$ ms, after three newer observations have arrived. RESYNC updates this stale latent with the new observations to approximate the latent of the current window, and the teacher's unchanged prediction head (H) turns it into the prediction for $t=4$ ms. The body then starts on the newest window, and every arrival until it finishes is served in the same way from the latest anchor.}
\label{fig:overview}
\end{figure}

We propose \textbf{RESYNC}, a lightweight updater that reconstructs the teacher's current latent from the latest stale one. We split the frozen teacher into an expensive body and a lightweight head, $f_T=H_T\circ B_T$. The body runs asynchronously and produces exact latent anchors $z_a=B_T(w_a)$ for windows $w_a$. At every arrival, RESYNC takes the newest anchor and the $n$ observations received since, predicts the displacement the body would have undergone over them, and applies the unchanged head. Each estimate starts from the newest exact anchor rather than from an earlier estimate, so update errors are not carried forward and one updater serves every anchor age. The frozen body supplies the training targets $z_{a+n}-z_a$, so fitting and selecting the updater needs no task labels. Figure~\ref{fig:overview} illustrates one arrival.

We evaluate prediction quality at fixed anchor ages and deadline performance in live asynchronous replay on one real and two synthetic streams and five architecture families. In the twelve pairs with a material teacher advantage, RESYNC beats the best compact or distilled model at anchor ages of 4--12 observations. At 1 ms arrivals, no synchronous teacher prediction meets the deadline, while RESYNC serves at least 99.4\% of arrivals within the interval, around the same speed as or faster than the compact model. Direct updates avoid the error accumulation seen in a matched recursive updater at longer gaps.

Our contributions are as follows:
\begin{itemize}
    \item We formulate latent synchronization for streams that require one prediction per observation from a teacher slower than the arrival rate.
    \item We introduce a direct, gap-conditioned updater that uses the newest exact anchor and the complete observed suffix, reuses the frozen head and trains without task labels.
    \item We evaluate it under live asynchronous execution on one real and two synthetic streams and five architecture families, against stale, compact, distilled, recursive and incremental alternatives, measuring both deadline performance and resource use.
\end{itemize}

\section{Related Work}
\label{sec:related-work}

\paragraph{Streaming latency and asynchronous prediction.}
Streaming perception evaluates the output actually available at a query time and lets models skip inputs while busy~\citep{li-2020-towards}. Slow--fast systems combine expensive image or event processing with cheap updates~\citep{gehrig-2024-lowlatency,chu-2026-bridging}; the asynchronous tracker of Chu et al.\ updates shared backbone features from new events and reuses the base model's head. RESYNC therefore does not claim novelty for slow--fast execution or head reuse. Its service contract requires a distinct prediction for every observation, and its updater estimates the \emph{current latent of the same frozen window model} directly from an exact anchor and the observations since.

\paragraph{Feature reuse and temporal distillation.}
Keyframe methods reuse costly representations through multirate execution, flow propagation, cached regions, or fusion with a current-frame branch~\citep{shelhamer-2016-clockwork,zhu-2017-deep,li-2018-lowlatency,xu-2018-deepcache,jain-2019-accel}. Closest is Delta Distillation, whose cheap students learn differences in teacher activations and then update features iteratively after an exact keyframe~\citep{habibian-2022-delta}; ResidualViT and StreamDEQ also propagate earlier computation~\citep{soldan-2025-residualvit,ertenli-2022-streamdeq}. RESYNC instead learns one variable-gap displacement at the latent--head interface from the newest exact anchor and the \emph{entire observed suffix}, and never uses an approximate latent as the reference for the next prediction.

\paragraph{Incremental inference and fast alternatives.}
Architecture-specific incremental execution avoids full recomputation, exactly or by ignoring small changes~\citep{cavigelli-2017-cbinfer,parger-2022-deltacnn,hedegaardmorsing-2023-continual,kechris-2025-dont}; where it meets the deadline, learned synchronization is unnecessary. Compact and early-exit models reduce current-input cost~\citep{teerapittayanon-2016-branchynet,huang-2018-multiscale}, and cached-summary models combine an expensive history representation with recent events~\citep{alattrach-2026-cached}. Teacher-feature supervision is established~\citep{romero-2015-fitnets,habibian-2022-delta,soldan-2025-residualvit}, so label-free fitting is not new in itself. Our contribution is the direct, gap-conditioned reconstruction rule and its evaluation under a per-observation service contract.

\section{Problem formulation}
\label{sec:problem-formulation}

\paragraph{Service contract.}
Observations $x_t\in\mathbb{R}^{F}$ arrive at wall-clock times $A_t$, and each defines a classification request with target $y_t\in\{1,\ldots,K\}$ and completion time $C_t$. The system owes one prediction for every $t$, with reaction latency $\ell_t=C_t-A_t$ below a deadline $D$. Dropping or coalescing requests reduces queuing but does not satisfy this contract. For a regular stream with interval $\Delta=A_t-A_{t-1}$, a serial predictor whose service time exceeds $\Delta$ falls ever further behind.

\paragraph{Frozen teacher.}
Let $w_t=(x_{t-L+1},\ldots,x_t)$ be the length-$L$ window available after observation $t$. We decompose a pretrained teacher into a body and a prediction head,
\begin{equation}
    f_T=H_T\circ B_T,\qquad
    z_t=B_T(w_t),\qquad
    p_T(y\mid w_t)=\operatorname{softmax}(H_T(z_t)).
    \label{eq:teacher-decomposition}
\end{equation}
Both remain frozen. The split is chosen so that $H_T$ is inexpensive; the fresh latent $z_t$ is what RESYNC aims to recover, but computing it at every arrival is too slow.

\paragraph{Available asynchronous state.}
Suppose a body job evaluated window $w_a$ and produced the exact anchor $z_a=B_T(w_a)$ at time $c_a$. At arrival $A_t$ the newest available anchor, its age and the observations since are
\begin{equation}
    a(t)=\max\{a:c_a\leq A_t\},\qquad
    n_t=t-a(t),\qquad
    d_{a(t),n_t}=(x_{a(t)+1},\ldots,x_t).
    \label{eq:available-anchor}
\end{equation}
The online system thus has an exact but stale latent and every observation since its window, while the current latent $z_t$ is available only as an offline training target. Synchronization is worthwhile when the teacher is materially better than every model that meets the deadline, its stale predictions lose that advantage, no suitable exact incremental implementation exists, and the complete foreground path keeps up with the arrivals while background compute is available for anchor refreshes.

\section{The RESYNC Method}
\label{sec:method}

\subsection{Direct latent synchronization}
For an exact anchor $z_a$, observed suffix $d_{a,n}$ and gap $n$, RESYNC predicts a latent displacement and applies the frozen head:
\begin{equation}
    \widehat\delta_{a,n}=f_R(z_a,d_{a,n},n),\qquad
    \widehat z^R_{a+n}=z_a+\widehat\delta_{a,n},\qquad
    \widehat p^R_{a+n}=\operatorname{softmax}(H_T(\widehat z^R_{a+n})).
    \label{eq:resync-method}
\end{equation}
Every estimate starts from the newest exact anchor, and $\widehat z^R_{a+n}$ is never used as the anchor for the next prediction. Earlier approximation errors are therefore not carried into later states, and one updater handles variable body latency through the explicit gap input. The updater encodes the causal suffix with a one-layer GRU and combines it with the normalized anchor and a learned gap embedding:
\begin{align}
    h_{a,n} &= \operatorname{GRU}_{\theta}(d_{a,n}), \\
    r_{a,n} &= W_2\,\operatorname{GELU}\!\left(
        W_1[\operatorname{LN}(z_a);h_{a,n};e(n)]+b_1\right)+b_2, \\
    \widehat\delta_{a,n} &= \boldsymbol\alpha_n\odot r_{a,n},
    \qquad \boldsymbol\alpha_n=\operatorname{sigmoid}(\boldsymbol q_n),\quad \boldsymbol q_n\in\mathbb R^{\dim(z_a)}.
    \label{eq:updater-architecture}
\end{align}
The learned gate damps each latent dimension separately at each gap, and the residual form contains the stale baseline as the zero-update solution. The supported gaps are $1\leq n\leq n_{\max}$; an older anchor reuses the embedding and gate of $n_{\max}$.

\subsection{Label-free updater training}
Training examples $(z_a,d_{a,n},n,z_{a+n})$ are built from an unlabeled stream by evaluating the frozen body at two endpoints. We sample $n$ uniformly from $1,\ldots,n_{\max}$ and minimize normalized latent reconstruction error plus a KL term between the frozen head's predictions from the fresh and synchronized latents:
\begin{equation}
    \mathcal L_R(\theta)=
    \mathbb E_{a,n}\left[
        \frac{\|\widehat z^R_{a+n}-z_{a+n}\|_2^2}
        {\|z_{a+n}\|_2^2+\epsilon}
        +\operatorname{KL}\!\left(p_T(\cdot\mid w_{a+n})\,\|\,\widehat p^R_{a+n}\right)
    \right],
    \label{eq:resync-loss}
\end{equation}
Only the updater is optimized, and checkpoints are selected by the same combined objective on validation examples. Fitting and selecting the updater thus need no task labels; the teacher and the supervised baselines are trained with labels, and labels are used for evaluation.

\subsection{Incremental asynchronous execution}
Because the GRU is causal, its state is advanced once per observation, $h_{a,n}=g_{\theta}(x_{a+n},h_{a,n-1})$ with $h_{a,0}=0$, so the work per arrival is constant in $n$ and $L$. One asynchronous worker runs the body. When it launches on window $w_a$, RESYNC starts a candidate encoder state and advances it with every later observation; once $z_a$ completes, the candidate already summarizes all observations since $a$ and becomes the active state. At each arrival (Figure~\ref{fig:overview}) the system installs any completed anchor, advances the encoder, applies Equation~\ref{eq:updater-architecture} and the head, and relaunches the body on the newest window when the worker is free. Anchor-dependent terms of the first MLP layer are computed once per anchor. The body runs outside the deadline-critical path. Because it may contend with the foreground path, we measure the complete system under concurrent replay.

\section{Experimental Design}
\label{sec:experimental-design}

\paragraph{Datasets.} We evaluate one real and two generated streams, each replayed in order with a causal window of $L=128$ observations that never crosses a day or run boundary. FI-2010~\citep{ntakaris-2018-benchmark} provides limit-order-book events from ten trading days; we predict the three-class mid-price movement at the 10-event horizon from 40 normalized order-book features, with seven development days (the last 20\% for validation) and three test days. Real streams give little control over how much long-context computation a label needs, so we add two generated streams in which it is designed in. In Synthetic-1, a rule revealed by two cue packets selects which component of two hidden oscillators defines the label; in Synthetic-2, the label is a delayed and gated combination of autoregressive sources, encoded in three single-event cues. Synthetic-2 deliberately lets some label-relevant cues or lagged signals fall outside the 128-event input window, adding uncertainty to the task. Both have three balanced classes and 18 runs of 50{,}000 events, split into 12 training, 3 validation and 3 test runs (Appendix~\ref{app:synthetic}).

\paragraph{Models.} To test that RESYNC needs only a body--head cut, we use five families covering attention, convolution with recurrence, pure convolution, state-space models and all-MLP mixing: a Transformer encoder, DeepConvLSTM~\citep{ordonez-2016-deep}, InceptionTime~\citep{ismailfawaz-2020-inceptiontime}, Mamba-2~\citep{dao-2024-transformers} and TSMixer~\citep{chen-2023-tsmixer}, all with the same lightweight head. For each dataset--architecture pair we fix a teacher (0.24--31.9M parameters) and a compact model (0.8k--76k) of the same family, chosen from six sizes so that the teacher is clearly slower and more accurate. The distilled compact model has the compact size and is trained against the teacher's logits. The RESYNC updater is the same for every pair: a one-layer GRU with 128 units, a 16-dimensional gap embedding and a 512-unit MLP (0.20--0.68M parameters), trained with gaps drawn uniformly from $1\leq n\leq n_{\max}=32$. It is larger than the compact models, which are compared with it at equal deadline, not equal size. All models use one training seed (Appendix~\ref{app:models}).

\paragraph{Live replay.} For every pair we replay the first 25{,}000 test events at fixed intervals $\Delta\in\{1,1.5,2,2.5,3\}$ ms with batch size one, three times in shuffled method order, on one NVIDIA GH200 node. The synchronous teacher runs on the GPU. The compact model and the RESYNC updater and head run on one CPU thread, while a worker thread launches body jobs on the GPU, each on the newest window. An arrival that is already one interval late when served is dropped. We report coverage, the fraction of arrivals that receive their own prediction, and deadline misses for $D=\Delta$. The main text uses $\Delta=1$ ms; Appendix~\ref{app:intervals} reports all intervals and Appendix~\ref{app:cost} runtime and resource measurements.

\paragraph{Comparisons.} Within each pair we compare RESYNC with the fresh synchronous teacher, the stale teacher (the head applied to the anchor $n$ observations old), and the compact and distilled models; the better of the latter two is the \emph{best fast model}. This name denotes the compact-size quality reference; the three Mamba-2 compact models do not meet the 1 ms deadline, as the live replay shows. A matched recursive updater tests the iterative-update mechanism of Delta Distillation~\citep{habibian-2022-delta} at the RESYNC split: the same module and recipe, trained on exact one-step pairs $(z_{t-1},x_t)\mapsto z_t$ and chained $n$ times from the anchor (Appendix~\ref{app:recursive}). We also compare with incremental inference by a Continual Transformer~\citep{hedegaard-2023-continual} and a recurrent Mamba-2 (Appendix~\ref{app:streaming}).

\paragraph{Quality measurements.} We measure macro-F1 on the same held-out targets at fixed anchor gaps $n\in\{1,4,8,12,16,18,22,26,30,32\}$, separately from the live replay. Paired 95\% block-bootstrap intervals over test days or runs assess the difference from the best fast model (Appendix~\ref{app:quality-details}). The live anchor ages show which fixed-gap quality measurements are most relevant to the deadline setting. We summarize the share of the teacher's gain captured by RESYNC only where the teacher leads the best fast model by at least 0.04 macro-F1.

\section{Results}

\subsection{Quality across pairs and anchor ages}

\begin{figure}[t]
\centering
\includegraphics[width=\textwidth]{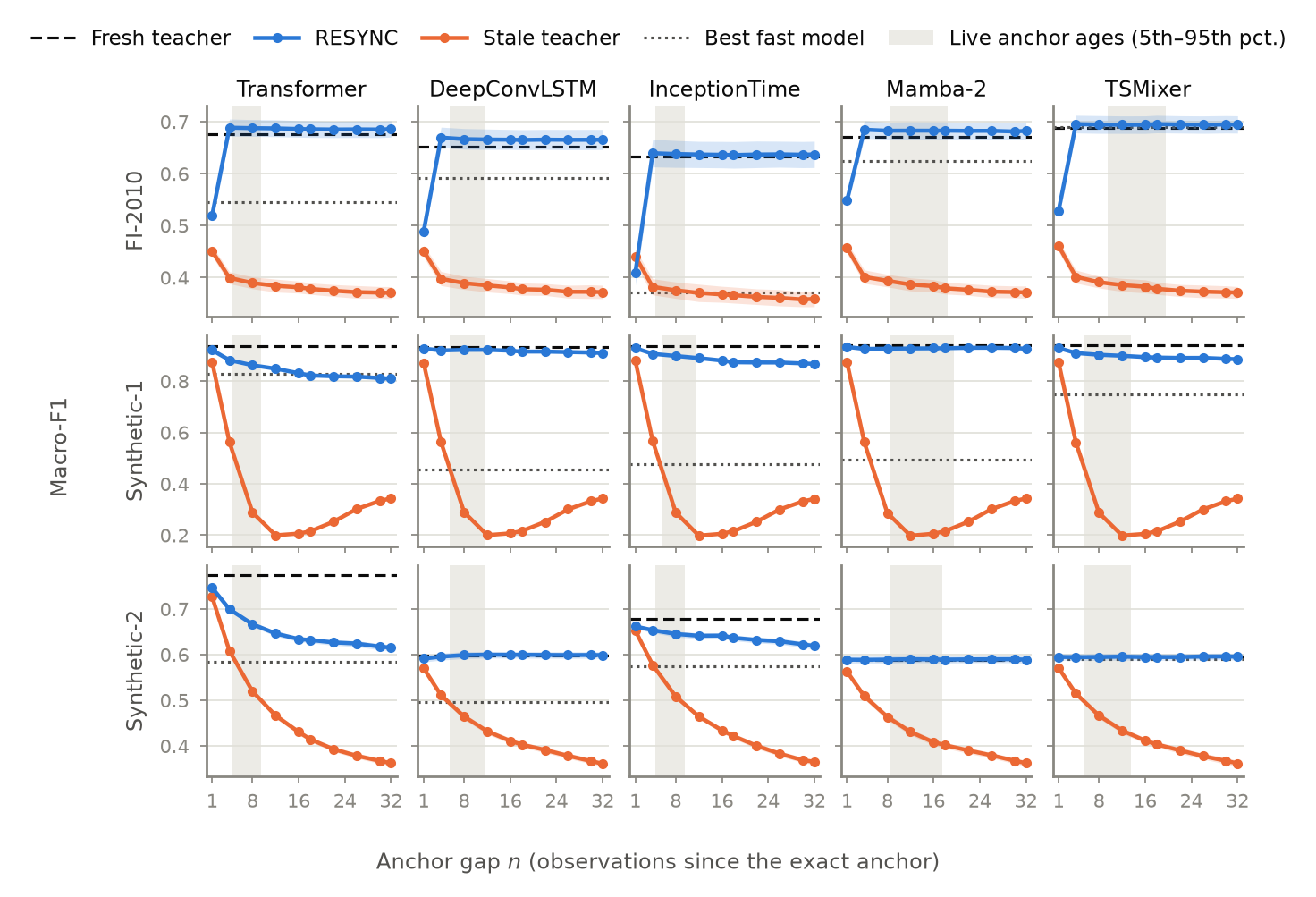}
\caption{Test macro-F1 as the anchor ages, for every dataset--architecture pair. RESYNC and the stale teacher are shown with paired 95\% block-bootstrap intervals. The fresh teacher and the best fast model (the better of the compact and distilled models) do not depend on $n$. The shaded band marks the 5th--95th percentile of the anchor ages RESYNC saw in live replay (Table~\ref{tab:live}).}
\label{fig:gap}
\end{figure}

Table~\ref{tab:quality} and Figure~\ref{fig:gap} show the quality result. In twelve of the fifteen pairs, the fresh teacher leads the best fast model by 0.05--0.48 macro-F1; these are the \emph{regime pairs} that RESYNC targets. Their stale teacher predictions lose most of that advantage within a few observations. RESYNC beats the best fast model with a paired 95\% interval above zero in all twelve regime pairs at the tested gaps $n=4,8,12$, and in eleven at every older tested gap through $n=32$. The exception is the Synthetic-1 Transformer: its difference from the distilled compact model is not distinguishable from zero at $n=16$, and the distilled model is ahead from $n=18$.

The amount preserved depends on the stream and the anchor age. Across $n=4$--32, RESYNC captures all of the teacher's gain in the four FI-2010 regime pairs, a median 0.87 on Synthetic-1 and 0.61 on Synthetic-2 (Table~\ref{tab:quality}, ``Gain capt.''). Its FI-2010 macro-F1 changes by at most 0.005 from $n=4$ to $n=32$. On the synthetic streams the gain captured generally declines with age: at $n=32$ it is 0.72--0.97 in four Synthetic-1 pairs, but only 0.16 and 0.43 for the Synthetic-2 Transformer and InceptionTime. Across all twelve regime pairs, the median gain captured is 0.96--0.97 at each tested gap, while the mean is 0.74--0.88; the lower mean reflects the weaker pairs that the median hides. RESYNC remains a median 0.012--0.018 macro-F1 below the fresh teacher at each gap.

\begin{table}[t]
\caption{Test macro-F1 of the fresh teacher, the compact model, the distilled compact model (KD), the stale teacher and RESYNC at anchor gaps $n=8$ and $n=32$. Gain captured is the share of the teacher's gain over the better fast model that RESYNC captures, $(\mathrm{F1}_{\mathrm{RESYNC}}-\mathrm{F1}_{\mathrm{fast}})/(\mathrm{F1}_{\mathrm{teacher}}-\mathrm{F1}_{\mathrm{fast}})$, capped at 1 (--: teacher's gain below 0.04). $\Delta$ best fast model is RESYNC minus the better fast model, with a paired 95\% block-bootstrap interval.}
\label{tab:quality}
\centering
\resizebox{\textwidth}{!}{%
\begin{tabular}{llccccccccc}
\toprule
 & & & & & \multicolumn{4}{c}{$n=8$} & \multicolumn{2}{c}{$n=32$} \\
\cmidrule(lr){6-9}\cmidrule(lr){10-11}
Dataset & Architecture & Teacher & Compact & KD & Stale & RESYNC & Gain capt. & $\Delta$ best fast model & RESYNC & Gain capt. \\
\midrule
FI-2010 & Transformer & 0.676 & 0.545 & 0.543 & 0.390 & 0.688 & 1.00 & +0.143 [+0.128, +0.160] & 0.686 & 1.00 \\
 & DeepConvLSTM & 0.652 & 0.592 & 0.575 & 0.389 & 0.667 & 1.00 & +0.075 [+0.063, +0.087] & 0.665 & 1.00 \\
 & InceptionTime & 0.633 & 0.372 & 0.357 & 0.376 & 0.638 & 1.00 & +0.266 [+0.232, +0.304] & 0.637 & 1.00 \\
 & Mamba-2 & 0.670 & 0.529 & 0.625 & 0.394 & 0.683 & 1.00 & +0.058 [+0.054, +0.062] & 0.683 & 1.00 \\
 & TSMixer & 0.688 & 0.661 & 0.690 & 0.391 & 0.695 & -- & +0.004 [+0.002, +0.007] & 0.695 & -- \\
\midrule
Synthetic-1 & Transformer & 0.937 & 0.801 & 0.828 & 0.289 & 0.862 & 0.31 & +0.034 [+0.031, +0.036] & 0.810 & -0.17 \\
 & DeepConvLSTM & 0.932 & 0.439 & 0.456 & 0.290 & 0.922 & 0.98 & +0.466 [+0.461, +0.472] & 0.909 & 0.95 \\
 & InceptionTime & 0.935 & 0.473 & 0.475 & 0.288 & 0.898 & 0.92 & +0.423 [+0.416, +0.429] & 0.867 & 0.85 \\
 & Mamba-2 & 0.940 & 0.493 & 0.477 & 0.287 & 0.927 & 0.97 & +0.434 [+0.432, +0.436] & 0.926 & 0.97 \\
 & TSMixer & 0.939 & 0.725 & 0.749 & 0.287 & 0.903 & 0.81 & +0.154 [+0.150, +0.157] & 0.886 & 0.72 \\
\midrule
Synthetic-2 & Transformer & 0.774 & 0.584 & 0.583 & 0.520 & 0.667 & 0.43 & +0.083 [+0.077, +0.088] & 0.615 & 0.16 \\
 & DeepConvLSTM & 0.597 & 0.497 & 0.496 & 0.464 & 0.599 & 1.00 & +0.103 [+0.096, +0.109] & 0.599 & 1.00 \\
 & InceptionTime & 0.678 & 0.575 & 0.575 & 0.508 & 0.645 & 0.68 & +0.070 [+0.064, +0.076] & 0.619 & 0.43 \\
 & Mamba-2 & 0.588 & 0.585 & 0.587 & 0.464 & 0.589 & -- & +0.002 [-0.001, +0.004] & 0.589 & -- \\
 & TSMixer & 0.597 & 0.583 & 0.590 & 0.467 & 0.595 & -- & +0.005 [+0.002, +0.008] & 0.596 & -- \\
\bottomrule
\end{tabular}
}
\end{table}

In the other three pairs---FI-2010 TSMixer and Synthetic-2 Mamba-2 and TSMixer---the teacher is at most 0.007 macro-F1 better than the fast models, leaving little gain for RESYNC to preserve. At $n=1$, RESYNC is weak on FI-2010 (0.65--0.82 of the teacher's macro-F1); this age hardly occurs in live replay. RESYNC also slightly exceeds its FI-2010 teachers, which we trace to a shift in class frequencies between training and test days rather than to a better representation (Appendix~\ref{app:above-teacher}). Appendices~\ref{app:quality-details} and~\ref{app:per-gap} give every pair at every gap.

\subsection{Live asynchronous replay}
\label{sec:results-live}

\begin{table}[t]
\caption{Live asynchronous replay at $\Delta=1$ ms: median arrival-to-output latency (ms) and coverage (the fraction of arrivals that received their own prediction) for the synchronous teacher, the compact model and RESYNC, and RESYNC's p95 latency. Anchor age is the median [p95, max] number of observations since RESYNC's anchor. Speed-up is the teacher's median service time over RESYNC's.}
\label{tab:live}
\centering
\resizebox{\textwidth}{!}{%
\begin{tabular}{llccccccccc}
\toprule
 & & \multicolumn{2}{c}{Teacher} & \multicolumn{2}{c}{Compact} & \multicolumn{3}{c}{RESYNC} & & \\
\cmidrule(lr){3-4}\cmidrule(lr){5-6}\cmidrule(lr){7-9}
Dataset & Architecture & p50 & Coverage & p50 & Coverage & p50 & p95 & Coverage & Anchor age & Speed-up \\
\midrule
FI-2010 & Transformer & 4.61 & 0.240 & 0.80 & 0.999 & 0.46 & 0.54 & 0.998 & 7 [9, 73] & 10.7$\times$ \\
 & DeepConvLSTM & 5.77 & 0.187 & 0.48 & 1.000 & 0.37 & 0.52 & 1.000 & 9 [11, 19] & 17.3$\times$ \\
 & InceptionTime & 4.57 & 0.242 & 0.65 & 1.000 & 0.59 & 0.64 & 0.999 & 7 [9, 37] & 7.9$\times$ \\
 & Mamba-2 & 7.39 & 0.143 & 1.60 & 0.902 & 0.59 & 0.66 & 0.999 & 14 [18, 41] & 13.4$\times$ \\
 & TSMixer & 7.76 & 0.136 & 0.90 & 0.999 & 0.62 & 0.68 & 1.000 & 15 [19, 31] & 13.4$\times$ \\
\midrule
Synthetic-1 & Transformer & 4.49 & 0.248 & 0.68 & 1.000 & 0.49 & 0.55 & 0.999 & 7 [9, 80] & 9.3$\times$ \\
 & DeepConvLSTM & 5.56 & 0.195 & 0.46 & 1.000 & 0.35 & 0.51 & 0.999 & 8 [11, 24] & 17.7$\times$ \\
 & InceptionTime & 5.74 & 0.187 & 0.80 & 0.999 & 0.62 & 0.68 & 0.998 & 9 [11, 25] & 9.6$\times$ \\
 & Mamba-2 & 7.06 & 0.135 & 1.56 & 0.930 & 0.57 & 0.64 & 0.998 & 13 [19, 57] & 13.2$\times$ \\
 & TSMixer & 5.07 & 0.216 & 0.50 & 1.000 & 0.56 & 0.62 & 0.999 & 9 [13, 44] & 9.5$\times$ \\
\midrule
Synthetic-2 & Transformer & 4.62 & 0.239 & 0.76 & 0.999 & 0.50 & 0.55 & 1.000 & 7 [9, 12] & 9.7$\times$ \\
 & DeepConvLSTM & 5.67 & 0.190 & 0.47 & 0.999 & 0.37 & 0.53 & 0.999 & 9 [11, 22] & 17.2$\times$ \\
 & InceptionTime & 4.76 & 0.231 & 0.79 & 0.997 & 0.56 & 0.62 & 0.999 & 7 [9, 21] & 8.6$\times$ \\
 & Mamba-2 & 6.43 & 0.165 & 1.24 & 0.969 & 0.58 & 0.65 & 0.997 & 13 [17, 21] & 11.8$\times$ \\
 & TSMixer & 5.00 & 0.219 & 0.51 & 1.000 & 0.57 & 0.64 & 0.997 & 9 [13, 42] & 9.0$\times$ \\
\bottomrule
\end{tabular}
}
\end{table}

Table~\ref{tab:live} tests the service contract at $\Delta=1$ ms. None of the fifteen synchronous teachers can sustain the arrival rate: only 14--25\% of arrivals receive a prediction, and none within one interval. Twelve compact models keep up, with median latencies of 0.46--0.90 ms; the three Mamba-2 compact models, at 1.24--1.60 ms, miss 92--100\% of the deadlines. RESYNC keeps up in every pair while the teacher runs in the background. Its median latency is 0.35--0.62 ms and its p99 at most 0.71 ms, below the interval in every pair. At most 0.57\% of arrivals miss the deadline; they are dropped during rare pauses of the replay process of up to 123 ms, which hit the compact models as well, rather than because of RESYNC's service time. RESYNC is faster than the compact model in thirteen pairs and at most 0.06 ms slower in the other two: around the same speed or faster overall. Its deadline-critical path is 7.9--17.7$\times$ faster than synchronous teacher service.

The anchor ages follow from the body's latency. At 1 ms their median is 7--15 observations and their 95th percentile at most 19, well inside the trained range; only 98 of 1.1 million predictions exceed $n_{\max}$, each just after a pause. Slower streams of 1.5--3 ms do not change the contrast between the methods but give younger anchors---at 3 ms, half of RESYNC's predictions use $n=2$ in eleven pairs---so we report 1 ms, the fastest tested stream (Appendix~\ref{app:intervals}).

\begin{table}[t]
\caption{The three systems at $\Delta=1$ ms, as median (range) over the fifteen pairs. Latency, coverage and missed deadlines ($D=\Delta$) come from the replay of Table~\ref{tab:live}; CPU, GPU and memory from a replay that runs each method in its own process (Appendix~\ref{app:cost}). CPU cores is process CPU time over wall time, GPU busy the share of the replay with a GPU call in flight, and memory the peak. The compact models' missed deadlines reach 100\% only for the three Mamba-2 compact models.}
\label{tab:system}
\centering
\small
\begin{tabular}{lccc}
\toprule
 & Synchronous teacher & Compact model & RESYNC \\
 & (GPU) & (CPU) & (CPU, body on GPU) \\
\midrule
Latency p50 (ms) & 5.56 (4.49--7.76) & 0.76 (0.46--1.60) & 0.56 (0.35--0.62) \\
Coverage (\%) & 19.5 (13.5--24.8) & 99.9 (90.2--100.0) & 99.9 (99.7--100.0) \\
Missed deadlines (\%) & 100.0 (100.0--100.0) & 0.2 (0.0--100.0) & 0.1 (0.0--0.6) \\
CPU cores & 1.00 (1.00--1.00) & 0.63 (0.40--0.99) & 1.34 (1.21--1.37) \\
GPU busy (\%) & 98 (98--99) & -- & 90 (85--93) \\
Peak host memory (GB) & 1.19 (1.04--1.33) & 0.66 (0.65--0.70) & 1.20 (1.06--1.34) \\
Peak GPU memory (MB) & 59 (36--172) & -- & 92 (45--190) \\
\bottomrule
\end{tabular}

\end{table}

Table~\ref{tab:system} puts the three systems side by side. RESYNC's per-arrival work does not depend on the anchor age: a controlled sweep times the complete tick at 0.22--0.37 ms on one CPU thread from $n=1$ to $n=80$, whereas re-encoding the observed suffix at every arrival would grow by 23--27~$\mu$s per observation. Ticks that install a new anchor keep a p99 latency of at most 0.74 ms. The worker relaunches the body every 5--10 arrivals, with a GPU call in flight for 85--93\% of the replay; Table~\ref{tab:system} and Appendix~\ref{app:cost} report the resource measurements.

\subsection{Direct re-synchronization versus recursive latent updates}
\label{sec:results-recursive}

\begin{figure}[t]
\centering
\includegraphics[width=\textwidth]{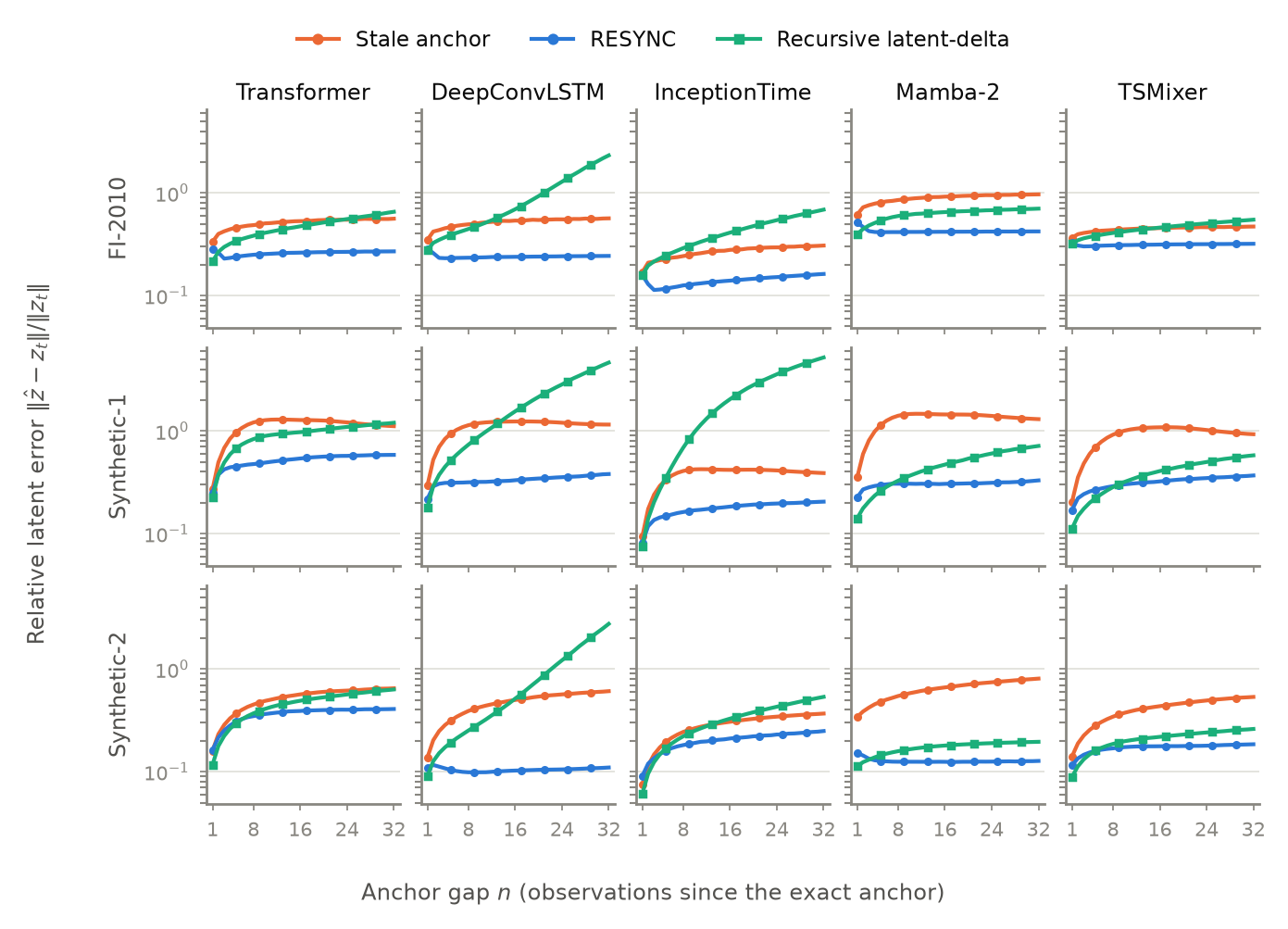}
\caption{Relative latent reconstruction error $\|\widehat z-z_t\|/\|z_t\|$ as the anchor ages, for the stale anchor, RESYNC and the matched recursive latent-delta updater. The recursive updater accumulates error; RESYNC's stays bounded.}
\label{fig:latent-error}
\end{figure}
\begin{table}[t]
\caption{Incremental inference at RESYNC's speed: median time per arrival (ms) and test macro-F1 of RESYNC on the teacher of each family, of the largest incremental model no slower than RESYNC, and of the next size up. RESYNC runs on one CPU thread. Each incremental model is timed as one step on one CPU thread, one step on the GPU and re-encoding the full window on the GPU, and shown at its fastest; $^{g}$: re-encoding on the GPU, otherwise the CPU step. RESYNC's F1 is at the evaluated gap nearest its median live anchor age. Appendix~\ref{app:streaming} includes the full grid.}
\label{tab:incremental-summary}
\centering
\small
\setlength{\tabcolsep}{4pt}
\begin{tabular}{llccccccc}
\toprule
 & & & & \multicolumn{4}{c}{Incremental} & \\
\cmidrule(lr){5-8}
 & & \multicolumn{2}{c}{RESYNC} & \multicolumn{2}{c}{$\leq$ RESYNC's time} & \multicolumn{2}{c}{Next size} & Teacher \\
\cmidrule(lr){3-4}\cmidrule(lr){5-6}\cmidrule(lr){7-8}
Dataset & Family & ms & F1 & ms & F1 & ms & F1 & F1 \\
\midrule
FI-2010 & Continual Transformer & 0.38 & 0.688 & 0.27 & 0.603 & 1.22$^{g}$ & 0.654 & 0.676 \\
 & Mamba-2, recurrent & 0.51 & 0.683 & 0.29 & 0.646 & 1.00 & 0.677 & 0.670 \\
\midrule
Synthetic-1 & Continual Transformer & 0.42 & 0.862 & 0.29 & 0.462 & 1.33$^{g}$ & 0.919 & 0.937 \\
 & Mamba-2, recurrent & 0.49 & 0.927 & 0.32 & 0.540 & 0.97 & 0.933 & 0.940 \\
\midrule
Synthetic-2 & Continual Transformer & 0.42 & 0.667 & 0.30 & 0.583 & 1.03$^{g}$ & 0.621 & 0.774 \\
 & Mamba-2, recurrent & 0.50 & 0.590 & 0.28 & 0.588 & 0.90 & 0.597 & 0.588 \\
\bottomrule
\end{tabular}

\end{table}

Figure~\ref{fig:latent-error} compares RESYNC with the matched recursive updater on all fifteen pairs (Appendix~\ref{app:recursive}). The recursive updater, trained only on one-step deltas, is better at $n=1$ in twelve pairs, and at $n=2$ and 3 the two are about even, with RESYNC ahead in nine. From $n=4$ onwards RESYNC is better in 13--15 pairs at every gap, and at $n=32$ its mean macro-F1 is 0.723 against 0.478 for the recursive updater and 0.360 for the stale anchor. The latent error shows why. RESYNC's error grows by a median factor of 1.5 from $n=1$ to $n=32$, because every estimate starts again from the exact anchor, whereas the recursive updater's grows by a factor of 5.2, because each step inherits the previous step's error. At $n=32$ the recursive updater's median relative error of 0.69 exceeds even the stale anchor's 0.61, against 0.27 for RESYNC. The label-free selection criterion already reveals the difference: the selected RESYNC checkpoints remove 0.77--0.98 of the stale validation loss, the recursive ones between $-3.37$ and 0.95.

\subsection{Incremental inference at RESYNC's speed}

Some architectures avoid re-encoding the window without a learned updater: a Continual Transformer~\citep{hedegaard-2023-continual} updates a shallow encoder exactly as observations arrive, and Mamba-2 can run as a recurrence. Table~\ref{tab:incremental-summary} compares each family with RESYNC on the teacher of the same family. To make the comparison fair to these models, we time each of them on one CPU thread and on the GPU, stepping or re-encoding the window, and report its fastest option. The available sizes bracket rather than match RESYNC's time: the largest incremental models no slower than RESYNC, all fastest on the CPU, trail it by 0.04--0.09 macro-F1 on FI-2010, by about 0.4 on Synthetic-1 and by 0.002--0.084 on Synthetic-2. One size up, at 1.8--3.2$\times$ RESYNC's time per arrival, they come close to RESYNC or pass it, except for the Continual Transformer on Synthetic-2. Their 0.90--1.33 ms per arrival does not fit 1 ms arrivals, however, even where the GPU is faster. The incremental models were replayed at 3 ms and scored at every test position, while RESYNC's timing uses 1 ms replay and its quality uses fixed-gap targets; recurrent Mamba-2 also only approximates the windowed model (Appendix~\ref{app:streaming}).

\section{Discussion and Limitations}
\label{sec:discussion}

RESYNC is intended for streams with background compute available but a tight deadline for each response. Its value depends on three conditions: the fresh teacher must improve on models that meet the deadline, stale anchors must lose that advantage, and the current latent must remain recoverable from an exact anchor and the observed suffix at the ages produced by the system. The first two conditions hold in twelve of our fifteen pairs; the third varies with the stream and anchor age. Quality changes little from $n=4$ to $n=32$ on FI-2010, but generally declines with age on the synthetic streams, and the Synthetic-1 Transformer falls below its distilled compact model from $n=18$. Teacher quality alone therefore does not establish the benefit of synchronization; fixed-gap quality must be considered alongside the live anchor ages.

Direct reconstruction has its clearest advantage over recursive updates at longer gaps: RESYNC is better in 13--15 pairs at every tested gap from $n=4$ to $32$, and the median live anchor ages at 1 ms are 7--15 observations. At $n=1$, the recursive updater is better in twelve pairs, though only one of about 1.1 million live predictions at 1 ms used $n\leq2$. Improving this short-gap case remains future work.

\paragraph{Deploying RESYNC.} The body's latency $T_{\mathrm{body}}$ and the arrival interval set the anchor ages, from about $T_{\mathrm{body}}/\Delta$ to $2T_{\mathrm{body}}/\Delta$ observations, and $n_{\max}$ should cover them with a margin (at 1 ms their 95th percentile was at most 19 for $n_{\max}=32$). The foreground path must meet the deadline while the body runs concurrently. The refresh schedule should be judged against the quality-versus-age curve: less frequent refreshes may be viable where quality changes little with age, as on FI-2010, but this scheduling choice was not tested.

\paragraph{Limitations.} The evaluation has one real stream, and the synthetic streams were built to require long-context computation, which may overstate how often a teacher advantage exists in practice. All models use one training seed, so the intervals reflect test-sample variation only. We study three-class classification with 128-observation windows, batch size one, fixed arrival intervals and one type of GPU node. Quality is measured at fixed gaps separately from service in live replay, rather than on the predictions delivered by that replay. The updater supports anchor ages up to $n_{\max}=32$, so slower bodies or faster streams would need a larger range. Its robustness to changing stream dynamics remains untested. RESYNC also needs a frozen teacher with an accessible latent.

\section{Conclusion}
\label{sec:conclusion}

RESYNC separates full-window teacher inference from the deadline-critical response: exact latents arrive asynchronously, and a direct updater estimates the current latent without carrying previous update errors forward. At the tested gaps $n=4,8,12$, RESYNC outperforms the better of the compact and distilled models in all twelve pairs with a material teacher advantage. In separate live replay at 1 ms, it serves at least 99.4\% of arrivals within the interval, while no synchronous teacher prediction meets the deadline. These results support latent synchronization for streams in which background computation is available and the teacher's latent remains recoverable over the anchor ages the system produces.

\section*{AI use statement}
In this study, we have used generative AI tools to support background research and identify previous related work in the domain. During prototyping and experimentation, we have used AI assistants to speed up coding and debugging. We have also used it for aiding in planning the high-level structure of individual sections, and proofreading and polishing our writing to improve the readability of this work. All text and code produced by AI is manually reviewed and re-written if necessary for correctness.

We have not used generative AI tools for developing conceptual frameworks, proposing or refining hypotheses, designing or providing feedback on research methodology or experiments, cleaning or reformatting datasets, supporting qualitative or thematic data analysis, or interpreting the results. AI-assistance in developing theoretical models, formulating mathematical claims, proving mathematical claims, writing proofs, assisting with translation are not applicable to this work.

We take responsibility for the full final content of this work.

\section*{Reproducibility statement}
The supplementary material of this study provides the code necessary for reproducing the experiments, including the five model families, the RESYNC updater, the incremental-inference baselines, the training of models, and the scripts for quality evaluation, live replay, runtime and resource measurements, and the recursive-updater study. The two synthetic datasets are generated from fixed seeds by the included generators, and FI-2010 is publicly available~\citep{ntakaris-2018-benchmark}. Model sizes, training settings and evaluation protocols are given in Appendices~\ref{app:models} and~\ref{app:quality-details}, and the timing measurements used one NVIDIA GH200 node (Section~\ref{sec:experimental-design}).

\bibliography{references}
\bibliographystyle{iclr2027_conference}


\appendix

\section{Synthetic stream generators}
\label{app:synthetic}
Real streams give little control over how much long-context computation a label requires, so it is hard to guarantee that a high-capacity teacher is materially better than a compact model. The two generated streams make labels depend on past cues and signals that a model must retrieve and combine with the current input. In both, the hidden score and rule state are absent from the inputs, and the three classes are balanced by thresholding a smooth hidden score.
\begin{itemize}
    \item \textbf{Synthetic-1} (14 channels) mixes three oscillators with irregular speeds into ten noisy sensor channels and adds two cue-value and two cue-marker channels. A rule, redrawn every 32--64 events, selects which sine or cosine component of the first two oscillators defines the score; the third oscillator is a distractor. Two short cue packets 16 events apart reveal the rule, so a model must retrieve both packets from its history and track the selected oscillator phase. Labels change every 8.7 events on average.
    \item \textbf{Synthetic-2} (20 channels) drives ten noisy sensor channels with three autoregressive latent sources and adds seven cue-value and three marker channels. Each 48--80-event segment draws a rule tier---the current value of one source (10\% of segments), a delayed combination of two sources (20\%), or a gated combination conditioned on a third, delayed source (70\%)---together with a source assignment, a sign and lags from $\{8,24,64\}$. The rule is encoded in three single-event cue packets and is not repeated between packets. Labels change every 4.9 events on average.
\end{itemize}
The Synthetic-2 generator's safety check used a 256-event window, whereas all evaluated models receive only 128 events. With lags up to 64 and sparse cue packets in segments up to 80 events long, some information used to form a label can lie outside the model's window. This is intentional: the shorter visible window adds uncertainty to the task.

Both streams are generated from fixed seeds, with scales set from the generation parameters rather than from any split. Table~\ref{tab:synthetic} lists the settings used.

\begin{table}[h]
\caption{Generator settings. Both streams have 12 training, 3 validation and 3 test runs of 50{,}000 events and three balanced classes.}
\label{tab:synthetic}
\centering
\small
\begin{tabular}{@{}p{3.4cm}p{4.5cm}p{5.1cm}@{}}
\toprule
Setting & Synthetic-1 & Synthetic-2 \\
\midrule
Channels & 10 sensor, 2 cue value, 2 marker & 10 sensor, 7 cue value, 3 marker \\
Hidden sources & 3 oscillators, mean periods 32, 46, 83 events & 3 AR(1) sources, memory 0.95 \\
Rule & 1 of 4 oscillator components & tier 0 / 1 / 2 with probability 0.1 / 0.2 / 0.7 \\
Segment length & 32--64 events & 48--80 events \\
Cue packets & 2 packets of 2 events, 16 events apart & 3 packets of 1 event, at offsets 0, 8, 16 \\
Lags & -- & $\{8, 24, 64\}$ \\
Source assignments & -- & 3 permutations \\
Rule codes held between packets & -- & no \\
Transition after the last cue & 6-event blend & 8-event blend \\
\raggedright Sensor noise / nuisance scale & 0.50 / 0.45 & 0.60 / 0.75 \\
\raggedright Mean events per label change (test) & 8.7 & 4.9 \\
Seed & 2027 & 2031 \\
\bottomrule
\end{tabular}
\end{table}

\section{Model configurations and training}
\label{app:models}
\paragraph{Architectures.} The Transformer encoder is adapted from the public implementation of karthik-1604\footnote{\url{https://github.com/karthik-1604/lob-midprice-forecasting/blob/master/models/transformer.py}}. All five families share the same head, LayerNorm, dropout and a linear layer, which holds well under 1\% of the teacher's parameters. The body's latent is the final-step vector for the Transformer, DeepConvLSTM and Mamba-2, the global average over the window for InceptionTime, and a learned temporal projection for TSMixer. Mamba-2 is implemented in plain PyTorch with the exact quadratic form of its state-space scan. Each family has six size variants. For every pair the teacher and compact sizes were chosen by testing several sizes until a clear divide was found, with the teacher substantially slower and more accurate than the compact model.

\paragraph{Sizes.} Table~\ref{tab:models} lists the teacher and compact sizes of every pair. The architecture sizes are: Transformer large, $d=256$, 16 layers, 8 heads, feed-forward 512, and small, $d=16$, 1 layer, 2 heads, feed-forward 32; DeepConvLSTM xlarge, 6 convolutions of 32 filters (kernel 6) and a 6-layer LSTM with 512 units, and mini, 1 convolution of 2 filters (kernel 1) and a 1-layer LSTM with 16 units; InceptionTime large, 9 modules with 64 filters per branch, kernels 39/19/9 and bottleneck 64, small, 3 modules with 4 filters (kernels 19/9/5), and mini, 3 modules with 2 filters (kernels 9/5/1); Mamba-2 base, $d=64$, 8 layers, state size 32, head dimension 16, and mini, $d=8$, 1 layer, state size 8, head dimension 4; TSMixer xlarge, $d=512$, 24 blocks, feed-forward 1024, large, $d=256$, 16 blocks, feed-forward 512, tiny, $d=32$, 2 blocks, and small, $d=16$, 1 block. The RESYNC updater has a 128-unit GRU, a 16-dimensional gap embedding and a 512-unit MLP.

\begin{table}[h]
\caption{Teacher and compact model of each pair, with parameter counts and the RESYNC updater's parameter count. $d$ is the latent dimension at the body--head cut.}
\label{tab:models}
\centering
\small
\begin{tabular}{llllrr}
\toprule
Dataset & Architecture & Teacher (params) & Compact (params) & $d$ & RESYNC params \\
\midrule
FI-2010 & Transformer & large (8.48M) & small (5.0k) & 256 & 411.4k \\
 & DeepConvLSTM & xlarge (11.67M) & mini (1.4k) & 512 & 682.8k \\
 & InceptionTime & large (2.89M) & mini (1.1k) & 256 & 411.4k \\
 & Mamba-2 & base (245.8k) & mini (1.1k) & 64 & 207.9k \\
 & TSMixer & xlarge (31.91M) & tiny (75.8k) & 512 & 682.8k \\
\midrule
Synthetic-1 & Transformer & large (8.47M) & small (4.6k) & 256 & 401.4k \\
 & DeepConvLSTM & xlarge (11.66M) & mini (1.4k) & 512 & 672.8k \\
 & InceptionTime & large (2.88M) & mini (0.8k) & 256 & 401.4k \\
 & Mamba-2 & base (244.2k) & mini (0.9k) & 64 & 197.9k \\
 & TSMixer & large (6.57M) & small (26.2k) & 256 & 401.4k \\
\midrule
Synthetic-2 & Transformer & large (8.47M) & small (4.7k) & 256 & 403.7k \\
 & DeepConvLSTM & xlarge (11.66M) & mini (1.4k) & 512 & 675.1k \\
 & InceptionTime & large (2.88M) & small (4.3k) & 256 & 403.7k \\
 & Mamba-2 & base (244.5k) & mini (1.0k) & 64 & 200.2k \\
 & TSMixer & large (6.57M) & small (26.3k) & 256 & 403.7k \\
\bottomrule
\end{tabular}

\end{table}

\paragraph{Training.} Teachers and compact models are trained with AdamW (learning rate $3\cdot10^{-4}$, weight decay $10^{-4}$), batch size 128, class-balanced sampling, bf16 mixed precision and gradient clipping at 1. Training runs for up to 50{,}000 steps with a 400-step warm-up and cosine decay, and the checkpoint with the best validation macro-F1, evaluated every 1{,}000 steps, is kept. Training stops after 10{,}000 steps without improvement. The knowledge-distilled compact model uses the same schedule with an equal mixture of cross-entropy and the temperature-2 distillation loss against the frozen teacher's logits. RESYNC is trained on cached latents of the frozen teacher for 100{,}000 steps with AdamW (learning rate $5\cdot10^{-4}$, 1{,}000 warm-up steps), batch size 128 and gaps drawn uniformly from $1\leq n\leq 32$. The KL term in Equation~\ref{eq:resync-loss} has weight 1; both terms are label-free. Normalizing the latent loss keeps examples with large latent norms from dominating it. The teacher is held in evaluation mode, and the RESYNC checkpoint is selected every 2{,}500 steps by the fraction of the stale combined validation loss it removes, without task labels. All results use one training seed.

\paragraph{Execution.} The incremental encoder state of Section~\ref{sec:method} is checked against the batched GRU output used during training, and the encoder step is included in the measured arrival-to-output latency. Approximate latents are never returned to the body or treated as exact anchors. Arrivals before the first exact anchor are counted as startup misses. The gap-embedding contribution to the first MLP layer is tabulated for $1,\ldots,n_{\max}$, and the anchor-dependent term is computed once when $z_a$ arrives; both preserve Equation~\ref{eq:updater-architecture} and only remove repeated work.

\section{Detailed quality results}
\label{app:quality-details}

\paragraph{Evaluation protocol.} Quality is evaluated on test positions only. A target $t$ is kept only if its anchors $t-n$ for every $n\leq32$ lie in the same test day or run, so every gap is scored on the same targets: 139{,}076 on FI-2010 and 149{,}492 on each synthetic stream. We report macro-F1 at $n\in\{1,4,8,12,16,18,22,26,30,32\}$, retention $\mathrm{F1}_{\mathrm{RESYNC}}/\mathrm{F1}_{\mathrm{teacher}}$, the share of the staleness loss recovered, $(\mathrm{F1}_{\mathrm{RESYNC}}-\mathrm{F1}_{\mathrm{stale}})/(\mathrm{F1}_{\mathrm{teacher}}-\mathrm{F1}_{\mathrm{stale}})$, and the advantage over the best fast model, whichever of the compact and distilled models scores higher by point estimate. The gain captured, the share of the teacher's gain over the best fast model that RESYNC captures, $(\mathrm{F1}_{\mathrm{RESYNC}}-\mathrm{F1}_{\mathrm{fast}})/(\mathrm{F1}_{\mathrm{teacher}}-\mathrm{F1}_{\mathrm{fast}})$, is 1 when RESYNC matches the teacher and 0 when it is no better than the best fast model. Because it is unstable where the teacher's lead is small, we summarize it only over the twelve regime pairs, whose teacher leads by at least 0.04. Retention, the gain captured and the share of the staleness loss recovered are capped at 1 in all summaries, so a RESYNC above its teacher counts as matching it. Intervals are paired 95\% block-bootstrap intervals: each test day or run is cut into contiguous blocks of 2{,}000 events (71 blocks on FI-2010, 75 on each synthetic stream), and all methods share the same 2{,}000 resamples. They capture test-sample variation, not training-seed variation.

\paragraph{Teacher advantage.} The fresh teacher is better than the best fast model by 0.05--0.26 macro-F1 for the FI-2010 Transformer, DeepConvLSTM, InceptionTime and Mamba-2, by 0.11--0.48 for all five Synthetic-1 pairs, and by 0.10--0.19 for the Synthetic-2 Transformer, DeepConvLSTM and InceptionTime. In the remaining three pairs the advantage is at most 0.007: on FI-2010 the distilled TSMixer matches its teacher, and the Synthetic-2 Mamba-2 and TSMixer teachers barely improve on their compact models. RESYNC's advantage over the fast models in these three pairs is 0.002--0.005 at $n=8$.

\paragraph{Stale and synchronized quality at $n=8$.} The stale teacher scores 0.38--0.39 on FI-2010, 0.29 on Synthetic-1 and 0.46--0.52 on Synthetic-2. RESYNC reaches 0.638--0.695, 0.862--0.927 and 0.589--0.667, a retention of 0.86--1.02. In the twelve regime pairs, its advantage over the best fast model at $n=8$ ranges from 0.034 to 0.466 macro-F1.

\paragraph{Across pairs.} RESYNC beats the best fast model with a paired interval above zero in fifteen pairs at $n=12$, fourteen at $n=4$, 8, 26 and 30, thirteen at $n=16$, 18, 22 and 32, and ten at $n=1$. The pairs that do not are Synthetic-2 Mamba-2 at seven of the ten gaps, the Synthetic-1 Transformer from $n=16$ and four FI-2010 pairs at $n=1$. Over the twelve regime pairs and $n\geq4$, RESYNC's capped retention has a median of 0.980--0.988 at each gap, it recovers a median 0.964--0.984 of the staleness loss, $(\mathrm{F1}_{\mathrm{RESYNC}}-\mathrm{F1}_{\mathrm{stale}})/(\mathrm{F1}_{\mathrm{teacher}}-\mathrm{F1}_{\mathrm{stale}})$, and it captures a median 0.96--0.97 and a mean 0.74--0.88 of the teacher's gain, with a minimum of 0.49 at $n=4$ falling to $-0.17$ at $n=32$ (the Synthetic-1 Transformer). Over the regime pairs of each dataset, the median gain captured is 1.00 on FI-2010, 0.87 on Synthetic-1 (from $-0.17$ to 0.98) and 0.61 on Synthetic-2 (from 0.16 to 1.00). Uncapped and over all fifteen pairs, the median retention would be 0.996--0.998; that figure is inflated by the pairs in which RESYNC matches or exceeds a teacher with little or no advantage, and we do not use it.

\paragraph{Synthetic-1 Transformer.} The distilled compact model reaches 0.828, and RESYNC's retention falls from 0.920 at $n=8$ to 0.864 at $n=32$. RESYNC is ahead of the distilled model with an interval above zero up to $n=12$ ($+0.020$ [$+0.018$, $+0.023$]). The two cannot be separated at $n=16$ ($+0.003$ [$-0.000$, $+0.006$]), and the distilled model is ahead from $n=18$ ($-0.006$ [$-0.009$, $-0.003$]) to $n=32$ ($-0.019$ [$-0.022$, $-0.015$]).

\paragraph{Bootstrap block length.} The median width of the 95\% interval of RESYNC minus the best fast model is 0.0102, 0.0106 and 0.0096 macro-F1 for blocks of 1{,}000, 2{,}000 and 5{,}000 events, so the intervals do not depend on the block length.

\paragraph{Pairs without a teacher advantage.} For Synthetic-2 Mamba-2, RESYNC's lead of about 0.002 over the distilled model cannot be separated from zero at seven of the ten gaps.

\paragraph{Anchor age.} At $n=32$, RESYNC retains 0.864--0.985 of the teacher on Synthetic-1, and 0.795 and 0.913 for the Synthetic-2 Transformer and InceptionTime; the other Synthetic-2 pairs stay at the teacher's level. The stale curve on Synthetic-1 is not monotone. It falls to about 0.20 at $n=12$--16 and partly recovers to 0.34 at $n=32$, because the periodic oscillators bring old predictions back into phase.

\paragraph{The first step.} At $n=1$, RESYNC behaves differently on FI-2010 than elsewhere. Its retention there is only 0.65--0.82, and for InceptionTime it is worse than the stale anchor, recovering $-0.161$ [$-0.217$, $-0.115$] of the staleness loss. For four of the five FI-2010 pairs it falls below the best fast model at $n=1$ by point estimate, which together with Synthetic-2 Mamba-2 is why the number of wins drops to ten of fifteen at that gap. On the synthetic streams, retention at $n=1$ is at least 0.96. The recursive study shows the same pattern: an updater trained only on one-step deltas is better than RESYNC at $n=1$ in all five FI-2010 pairs (Table~\ref{tab:recursive}). Uniform gap sampling gives $n=1$ only $1/32$ of the training examples, but we have not isolated the cause of its weaker FI-2010 performance. The weakness lies outside the live operating range: at every tested arrival interval, at most 0.004\% of the predictions of any pair use $n=1$, and at $\Delta=1$ ms only one of 1.1 million predictions uses $n\leq2$. At $\Delta=3$ ms, half of RESYNC's predictions use $n=2$ in eleven of the fifteen pairs (Appendix~\ref{app:intervals}); at $n=2$ the recursive study measures RESYNC within 0.010--0.030 of the teacher for all five FI-2010 architectures, and matching or exceeding it from $n=3$ onwards.

\paragraph{Retention above one.} RESYNC's retention slightly exceeds one in all five FI-2010 pairs and, by at most 0.003 macro-F1, in Synthetic-2 DeepConvLSTM and Mamba-2. On training positions RESYNC stays below each of these teachers (Appendix~\ref{app:above-teacher}), so we treat it as matching them.

\section{RESYNC above the fresh teacher}
\label{app:above-teacher}
On the FI-2010 test days, RESYNC scores slightly above the fresh teacher whose latent it approximates, for all five architectures (Table~\ref{tab:quality}). We checked three explanations; the analysis uses RESYNC at $n=8$ and the target positions of Table~\ref{tab:quality}.

\paragraph{Not an evaluation artefact.} RESYNC's estimate for target $t$ uses only the anchor latent at $t-n$ and the observations $x_{t-n+1},\ldots,x_t$, the same information the teacher's window ends with in isolated testing. On training positions RESYNC scores below the teacher for all five architectures (Table~\ref{tab:above-teacher}). A misalignment or leak would affect both splits.

\paragraph{Not temporal smoothing.} RESYNC changes its prediction between consecutive test events almost as often as the teacher: 20.5--28.2\% of steps against 20.8--30.0\% across the five architectures, and 27.1\% against 28.8\% for the Transformer. Causally smoothing the teacher's own predictions, by a majority vote over its last three, lowers its test macro-F1 from 0.633--0.688 to 0.485--0.544, and from 0.676 to 0.530 for the Transformer.

\paragraph{A shift in class frequencies.} The label shares (up/stationary/down) are 0.20/0.60/0.20 on the training days, 0.18/0.63/0.18 on validation and 0.15/0.71/0.14 on test. The teacher is trained with class-balanced sampling and over-predicts price moves on test: the Transformer predicts ``stationary'' for 65.6\% of test positions, against a true 70.8\%. RESYNC's imperfect latent estimates move some of the teacher's move predictions to ``stationary'' (68.5\%). The gain comes from exactly these changes (Table~\ref{tab:above-teacher}). Where the teacher predicts a move and RESYNC predicts ``stationary'', on 3--6\% of test positions, RESYNC is right 46--72\% of the time and the teacher 21--41\%. Where RESYNC moves away from ``stationary'', the teacher is right more often. Applying only RESYNC's changes to ``stationary'' to the teacher's predictions scores higher than RESYNC itself for the Transformer, Mamba-2 and TSMixer. On the training days, where moves are more frequent, the same changes raise the teacher's macro-F1 by at most 0.006, against up to 0.021 on test. RESYNC therefore acts as an unintended shift toward the majority class, and that shift happens to suit the test days.

\begin{table}[h]
\caption{FI-2010, RESYNC at $n=8$. Macro-F1 of the teacher (T) and RESYNC (R) on training and test positions. ``$\to$ stationary'': test positions where the teacher predicts a move and RESYNC predicts stationary, with the fraction of them each model gets right; ``stationary $\to$'' is the reverse. The last column is the teacher's macro-F1 with only the ``$\to$ stationary'' changes applied.}
\label{tab:above-teacher}
\centering
\resizebox{\textwidth}{!}{%
\begin{tabular}{lcccccccccc}
\toprule
 & \multicolumn{2}{c}{Train F1} & \multicolumn{2}{c}{Test F1} & \multicolumn{3}{c}{$\to$ stationary (test)} & \multicolumn{2}{c}{stationary $\to$ (test)} & \\
\cmidrule(lr){2-3}\cmidrule(lr){4-5}\cmidrule(lr){6-8}\cmidrule(lr){9-10}
Architecture & T & R & T & R & Share & R right & T right & R right & T right & T + $\to$ stationary \\
\midrule
Transformer & 0.747 & 0.720 & 0.676 & 0.688 & 0.063 & 0.72 & 0.21 & 0.25 & 0.64 & 0.697 \\
DeepConvLSTM & 0.693 & 0.690 & 0.652 & 0.667 & 0.048 & 0.69 & 0.21 & 0.28 & 0.60 & 0.666 \\
InceptionTime & 0.705 & 0.690 & 0.633 & 0.638 & 0.031 & 0.46 & 0.41 & 0.44 & 0.45 & 0.619 \\
Mamba-2 & 0.731 & 0.712 & 0.670 & 0.683 & 0.052 & 0.69 & 0.22 & 0.27 & 0.63 & 0.685 \\
TSMixer & 0.735 & 0.718 & 0.688 & 0.695 & 0.053 & 0.62 & 0.28 & 0.28 & 0.62 & 0.696 \\
\bottomrule
\end{tabular}
}
\end{table}

\paragraph{Synthetic-2 Mamba-2 and DeepConvLSTM.} Here the classes are balanced in every split, and the differences at $n=8$ are 0.001 and 0.002 macro-F1 (0.589 vs.\ 0.588 and 0.599 vs.\ 0.597). RESYNC changes 5.2\% and 6.7\% of the teachers' test predictions and is right on 43\% of them, against 41\% for the teachers. On training positions it is at or below the teachers (0.5948 vs.\ 0.5952 and 0.607 vs.\ 0.610). Only the DeepConvLSTM interval excludes one, and the intervals cover test blocks, not retraining, so we report parity.

\section{Per-gap results}
\label{app:per-gap}
Table~\ref{tab:per-gap-f1} gives RESYNC's test macro-F1 at every evaluated anchor gap, with the fresh teacher and the best fast model for reference. Tables~\ref{tab:per-gap-diff-a} and~\ref{tab:per-gap-diff-b} give the paired difference between RESYNC and the best fast model with its 95\% block-bootstrap interval. Appendix~\ref{app:intervals} summarizes the anchor ages RESYNC sees in live replay.

\begin{table}[h]
\caption{Test macro-F1 of RESYNC at every evaluated anchor gap $n$, with the fresh teacher and the best fast model (the better of the compact and distilled models), which do not depend on $n$.}
\label{tab:per-gap-f1}
\centering
\resizebox{\textwidth}{!}{%
\begin{tabular}{llcccccccccccc}
\toprule
 & & & & \multicolumn{10}{c}{RESYNC at anchor gap $n$} \\
\cmidrule(lr){5-14}
Dataset & Architecture & Teacher & Best fast & $1$ & $4$ & $8$ & $12$ & $16$ & $18$ & $22$ & $26$ & $30$ & $32$ \\
\midrule
FI-2010 & Transformer & 0.676 & 0.545 & 0.519 & 0.689 & 0.688 & 0.687 & 0.686 & 0.686 & 0.685 & 0.685 & 0.685 & 0.686 \\
 & DeepConvLSTM & 0.652 & 0.592 & 0.489 & 0.670 & 0.667 & 0.666 & 0.666 & 0.665 & 0.666 & 0.665 & 0.665 & 0.665 \\
 & InceptionTime & 0.633 & 0.372 & 0.410 & 0.640 & 0.638 & 0.637 & 0.636 & 0.636 & 0.637 & 0.637 & 0.637 & 0.637 \\
 & Mamba-2 & 0.670 & 0.625 & 0.548 & 0.685 & 0.683 & 0.683 & 0.683 & 0.683 & 0.683 & 0.683 & 0.681 & 0.683 \\
 & TSMixer & 0.688 & 0.690 & 0.529 & 0.696 & 0.695 & 0.694 & 0.694 & 0.696 & 0.695 & 0.694 & 0.695 & 0.695 \\
\midrule
Synthetic-1 & Transformer & 0.937 & 0.828 & 0.921 & 0.882 & 0.862 & 0.849 & 0.831 & 0.822 & 0.819 & 0.818 & 0.813 & 0.810 \\
 & DeepConvLSTM & 0.932 & 0.456 & 0.927 & 0.920 & 0.922 & 0.922 & 0.919 & 0.915 & 0.915 & 0.913 & 0.912 & 0.909 \\
 & InceptionTime & 0.935 & 0.475 & 0.929 & 0.907 & 0.898 & 0.890 & 0.880 & 0.874 & 0.873 & 0.873 & 0.869 & 0.867 \\
 & Mamba-2 & 0.940 & 0.493 & 0.933 & 0.926 & 0.927 & 0.927 & 0.928 & 0.928 & 0.930 & 0.930 & 0.929 & 0.926 \\
 & TSMixer & 0.939 & 0.749 & 0.929 & 0.910 & 0.903 & 0.899 & 0.894 & 0.892 & 0.891 & 0.891 & 0.887 & 0.886 \\
\midrule
Synthetic-2 & Transformer & 0.774 & 0.584 & 0.747 & 0.700 & 0.667 & 0.646 & 0.634 & 0.632 & 0.627 & 0.624 & 0.618 & 0.615 \\
 & DeepConvLSTM & 0.597 & 0.497 & 0.591 & 0.596 & 0.599 & 0.600 & 0.600 & 0.600 & 0.599 & 0.599 & 0.599 & 0.599 \\
 & InceptionTime & 0.678 & 0.575 & 0.662 & 0.654 & 0.645 & 0.641 & 0.642 & 0.637 & 0.632 & 0.629 & 0.623 & 0.619 \\
 & Mamba-2 & 0.588 & 0.587 & 0.589 & 0.589 & 0.589 & 0.590 & 0.589 & 0.589 & 0.590 & 0.590 & 0.590 & 0.589 \\
 & TSMixer & 0.597 & 0.590 & 0.595 & 0.595 & 0.595 & 0.596 & 0.595 & 0.595 & 0.595 & 0.596 & 0.596 & 0.596 \\
\bottomrule
\end{tabular}
}
\end{table}

\begin{table}[h]
\caption{RESYNC minus the best fast model at $n=1$--16, with paired 95\% block-bootstrap intervals.}
\label{tab:per-gap-diff-a}
\centering
\resizebox{\textwidth}{!}{%
\begin{tabular}{llccccc}
\toprule
Dataset & Architecture & $n=1$ & $n=4$ & $n=8$ & $n=12$ & $n=16$ \\
\midrule
FI-2010 & Transformer & -0.025 [-0.052, +0.003] & +0.144 [+0.129, +0.161] & +0.143 [+0.128, +0.160] & +0.143 [+0.127, +0.160] & +0.141 [+0.126, +0.159] \\
 & DeepConvLSTM & -0.103 [-0.115, -0.090] & +0.078 [+0.066, +0.089] & +0.075 [+0.063, +0.087] & +0.074 [+0.062, +0.086] & +0.074 [+0.063, +0.085] \\
 & InceptionTime & +0.038 [+0.023, +0.055] & +0.268 [+0.233, +0.307] & +0.266 [+0.232, +0.304] & +0.265 [+0.231, +0.304] & +0.265 [+0.231, +0.303] \\
 & Mamba-2 & -0.077 [-0.086, -0.066] & +0.060 [+0.056, +0.064] & +0.058 [+0.054, +0.062] & +0.058 [+0.054, +0.062] & +0.058 [+0.054, +0.062] \\
 & TSMixer & -0.161 [-0.177, -0.145] & +0.005 [+0.002, +0.008] & +0.004 [+0.002, +0.007] & +0.004 [+0.001, +0.007] & +0.004 [+0.001, +0.007] \\
\midrule
Synthetic-1 & Transformer & +0.093 [+0.090, +0.095] & +0.054 [+0.051, +0.056] & +0.034 [+0.031, +0.036] & +0.020 [+0.018, +0.023] & +0.003 [-0.000, +0.006] \\
 & DeepConvLSTM & +0.472 [+0.466, +0.477] & +0.464 [+0.459, +0.470] & +0.466 [+0.461, +0.472] & +0.467 [+0.461, +0.472] & +0.463 [+0.458, +0.469] \\
 & InceptionTime & +0.454 [+0.447, +0.460] & +0.431 [+0.425, +0.437] & +0.423 [+0.416, +0.429] & +0.415 [+0.408, +0.421] & +0.405 [+0.398, +0.412] \\
 & Mamba-2 & +0.440 [+0.438, +0.443] & +0.433 [+0.431, +0.435] & +0.434 [+0.432, +0.436] & +0.435 [+0.432, +0.437] & +0.435 [+0.433, +0.437] \\
 & TSMixer & +0.180 [+0.177, +0.184] & +0.161 [+0.157, +0.164] & +0.154 [+0.150, +0.157] & +0.150 [+0.146, +0.154] & +0.145 [+0.141, +0.149] \\
\midrule
Synthetic-2 & Transformer & +0.163 [+0.155, +0.171] & +0.116 [+0.109, +0.122] & +0.083 [+0.077, +0.088] & +0.062 [+0.057, +0.067] & +0.050 [+0.045, +0.055] \\
 & DeepConvLSTM & +0.095 [+0.088, +0.101] & +0.099 [+0.092, +0.106] & +0.103 [+0.096, +0.109] & +0.103 [+0.097, +0.110] & +0.103 [+0.097, +0.110] \\
 & InceptionTime & +0.087 [+0.080, +0.094] & +0.078 [+0.073, +0.084] & +0.070 [+0.064, +0.076] & +0.066 [+0.060, +0.073] & +0.066 [+0.060, +0.072] \\
 & Mamba-2 & +0.002 [-0.001, +0.004] & +0.001 [-0.001, +0.004] & +0.002 [-0.001, +0.004] & +0.003 [+0.000, +0.005] & +0.002 [-0.000, +0.005] \\
 & TSMixer & +0.005 [+0.002, +0.008] & +0.005 [+0.002, +0.007] & +0.005 [+0.002, +0.008] & +0.006 [+0.003, +0.009] & +0.005 [+0.002, +0.008] \\
\bottomrule
\end{tabular}
}
\end{table}

\begin{table}[h]
\caption{RESYNC minus the best fast model at $n=18$--32, with paired 95\% block-bootstrap intervals.}
\label{tab:per-gap-diff-b}
\centering
\resizebox{\textwidth}{!}{%
\begin{tabular}{llccccc}
\toprule
Dataset & Architecture & $n=18$ & $n=22$ & $n=26$ & $n=30$ & $n=32$ \\
\midrule
FI-2010 & Transformer & +0.141 [+0.125, +0.159] & +0.140 [+0.125, +0.158] & +0.141 [+0.125, +0.158] & +0.140 [+0.125, +0.158] & +0.141 [+0.125, +0.159] \\
 & DeepConvLSTM & +0.073 [+0.062, +0.085] & +0.074 [+0.063, +0.085] & +0.073 [+0.062, +0.085] & +0.074 [+0.062, +0.085] & +0.074 [+0.062, +0.085] \\
 & InceptionTime & +0.264 [+0.231, +0.302] & +0.265 [+0.231, +0.303] & +0.266 [+0.232, +0.303] & +0.265 [+0.231, +0.302] & +0.265 [+0.232, +0.303] \\
 & Mamba-2 & +0.058 [+0.054, +0.062] & +0.058 [+0.054, +0.062] & +0.058 [+0.054, +0.062] & +0.056 [+0.053, +0.060] & +0.058 [+0.054, +0.062] \\
 & TSMixer & +0.005 [+0.003, +0.008] & +0.005 [+0.002, +0.007] & +0.004 [+0.001, +0.007] & +0.005 [+0.002, +0.007] & +0.004 [+0.002, +0.007] \\
\midrule
Synthetic-1 & Transformer & -0.006 [-0.009, -0.003] & -0.009 [-0.013, -0.006] & -0.011 [-0.015, -0.007] & -0.016 [-0.020, -0.012] & -0.019 [-0.022, -0.015] \\
 & DeepConvLSTM & +0.460 [+0.454, +0.465] & +0.460 [+0.454, +0.465] & +0.458 [+0.452, +0.464] & +0.456 [+0.450, +0.462] & +0.454 [+0.448, +0.460] \\
 & InceptionTime & +0.399 [+0.391, +0.405] & +0.398 [+0.390, +0.405] & +0.398 [+0.390, +0.404] & +0.394 [+0.387, +0.401] & +0.392 [+0.385, +0.399] \\
 & Mamba-2 & +0.436 [+0.433, +0.438] & +0.437 [+0.435, +0.439] & +0.437 [+0.435, +0.439] & +0.436 [+0.434, +0.439] & +0.433 [+0.431, +0.436] \\
 & TSMixer & +0.143 [+0.139, +0.147] & +0.142 [+0.137, +0.146] & +0.142 [+0.138, +0.146] & +0.138 [+0.134, +0.142] & +0.137 [+0.133, +0.141] \\
\midrule
Synthetic-2 & Transformer & +0.048 [+0.043, +0.053] & +0.043 [+0.038, +0.047] & +0.040 [+0.036, +0.045] & +0.034 [+0.029, +0.038] & +0.031 [+0.027, +0.036] \\
 & DeepConvLSTM & +0.103 [+0.096, +0.110] & +0.103 [+0.096, +0.109] & +0.103 [+0.096, +0.109] & +0.103 [+0.096, +0.109] & +0.102 [+0.096, +0.109] \\
 & InceptionTime & +0.062 [+0.056, +0.068] & +0.057 [+0.051, +0.063] & +0.054 [+0.048, +0.060] & +0.047 [+0.041, +0.054] & +0.044 [+0.038, +0.050] \\
 & Mamba-2 & +0.002 [-0.001, +0.005] & +0.002 [-0.000, +0.005] & +0.003 [+0.000, +0.005] & +0.003 [+0.000, +0.005] & +0.002 [-0.001, +0.004] \\
 & TSMixer & +0.005 [+0.002, +0.008] & +0.005 [+0.002, +0.008] & +0.006 [+0.003, +0.009] & +0.006 [+0.003, +0.009] & +0.006 [+0.003, +0.009] \\
\bottomrule
\end{tabular}
}
\end{table}

\section{Live replay at every arrival interval}
\label{app:intervals}
Tables~\ref{tab:intervals-summary} and~\ref{tab:intervals} report the live replay of every pair at $\Delta=1$, 1.5, 2, 2.5 and 3 ms, with the same teacher, compact model and RESYNC updater. Each replay has 74{,}100 timed arrivals per method (three repetitions of 25{,}000 events minus the warm-up).

\paragraph{Measures.} An arrival is \emph{dropped} if it is already one interval late when it is served. It then receives no prediction, although RESYNC still folds its observation into the encoder state. It is \emph{late} if its prediction completes more than one interval after its arrival. It \emph{misses} the reference deadline $D=\Delta$ if it is dropped, late or arrives before the first exact anchor. After the warm-up no replay has startup arrivals, so every missing prediction is a dropped arrival. The anchor age is counted over RESYNC's completed predictions. A prediction is \emph{clamped} if its age exceeds $n_{\max}=32$; it then reuses the gap embedding and gate of $n_{\max}$. The longest stall is the longest run of consecutive dropped arrivals within one repetition, multiplied by $\Delta$.

\begin{table}[h]
\caption{Live replay per arrival interval, over the fifteen pairs. Teacher coverage: range over pairs; no teacher prediction completes within one interval at any $\Delta$. Compact: number of pairs whose compact model misses more than 5\% of arrivals. RESYNC service: median and maximum dropped share, maximum miss share, longest stall and maximum p99 latency over pairs. Anchor age: range of the per-pair medians, maximum p99, maximum age and median share of predictions with $n\leq2$. $n>n_{\max}$: pairs with any clamped prediction, the number of clamped predictions (of about 1.1 million per interval), with how many of them lie within 40 arrivals of a dropped arrival in parentheses, and the largest clamped share of a pair.}
\label{tab:intervals-summary}
\centering
\resizebox{\textwidth}{!}{%
\begin{tabular}{ccccccccccccccc}
\toprule
 & Teacher & Compact & \multicolumn{5}{c}{RESYNC service} & \multicolumn{4}{c}{RESYNC anchor age} & \multicolumn{3}{c}{RESYNC $n>n_{\max}$} \\
\cmidrule(lr){2-2}\cmidrule(lr){3-3}\cmidrule(lr){4-8}\cmidrule(lr){9-12}\cmidrule(lr){13-15}
$\Delta$ & Coverage & Pairs & \multicolumn{2}{c}{Dropped (\%)} & Miss (\%) & Stall & p99 & Median & p99 & Max & $n\leq2$ (\%) & Pairs & Count & Share (\%) \\
(ms) & range & $>5\%$ miss & median & max & max & max (ms) & max (ms) & range & max & max & median & & (after stall) & max \\
\midrule
1 & 0.14--0.25 & 3 & 0.11 & 0.33 & 0.57 & 123 & 0.71 & 7--15 & 20 & 80 & 0.00 & 7 & 98 (98) & 0.066 \\
1.5 & 0.20--0.39 & 0 & 0.06 & 0.11 & 0.14 & 90 & 0.72 & 4--9 & 13 & 50 & 0.00 & 6 & 26 (23) & 0.012 \\
2 & 0.29--0.52 & 0 & 0.02 & 0.07 & 0.08 & 84 & 0.73 & 4--6 & 9 & 41 & 0.00 & 3 & 5 (5) & 0.003 \\
2.5 & 0.34--0.65 & 0 & 0.00 & 0.07 & 0.09 & 80 & 0.72 & 3--5.5 & 7 & 31 & 1.07 & 0 & 0 (0) & 0.000 \\
3 & 0.43--0.78 & 0 & 0.02 & 0.06 & 0.10 & 75 & 0.71 & 3--4 & 5 & 28 & 49.90 & 0 & 0 (0) & 0.000 \\
\bottomrule
\end{tabular}
}
\end{table}

\paragraph{Service at $\Delta=1$ ms.} The synchronous teachers' median arrival-to-output latency is 4.5--7.8 ms, and every delivered teacher prediction arrives later than one interval after its observation. The twelve compact models that keep up have coverage of at least 0.997; the three Mamba-2 compact models drop 3--10\% of their arrivals. At least 99.67\% of arrivals receive their own RESYNC prediction, and RESYNC's p95 latency is at most 0.68 ms. Its per-arrival work is one GRU step and two small matrix products regardless of window length, so it is faster than the compact model in thirteen pairs and at most 0.06 ms slower in the other two.

\paragraph{Anchor ages at 1 ms and the choice of $\Delta$.} The anchor ages RESYNC sees follow from the teacher body's latency $T_{\mathrm{body}}$: each anchor is used from roughly $T_{\mathrm{body}}/\Delta$ to $2T_{\mathrm{body}}/\Delta$ observations after its window. At $\Delta=1$ ms the median is 7--9 observations for the Transformer, DeepConvLSTM, InceptionTime and synthetic-stream TSMixer teachers, and 13--15 for the Mamba-2 and FI-2010 TSMixer teachers. Slower streams do not change the contrast between the methods: no synchronous teacher delivers a single prediction within one interval at any of them, every compact model keeps up from 1.5 ms, and RESYNC misses at most 0.14\% of arrivals. At 3 ms, however, half of RESYNC's predictions in eleven pairs use $n=2$, a gap that the quality evaluation of Table~\ref{tab:quality} does not cover. We therefore report 1 ms in the main text, the fastest tested stream, at which RESYNC still meets the deadline for at least 99.4\% of arrivals in every pair and the anchor ages lie well inside the trained range.

\paragraph{Teacher.} No synchronous teacher meets a single deadline at any interval. Its coverage rises from 14--25\% at 1 ms to 43--78\% at 3 ms, roughly the interval over its service time. Because overdue arrivals are skipped rather than queued, coverage moves in steps of that ratio and need not rise monotonically with $\Delta$; for the FI-2010 Transformer it is 0.49 at 2 ms and 0.47 at 2.5 ms.

\paragraph{Compact models.} At 1 ms the three Mamba-2 compact models are slower than the interval, with median latencies of 1.24--1.60 ms, and miss 92--100\% of arrivals. From 1.5 ms every compact model keeps up. The largest miss share is 2.7\% (Synthetic-2 Mamba-2 at 1.5 ms), and from 2 ms it is at most 0.11\%.

\paragraph{RESYNC service.} RESYNC keeps up in every pair at every interval. Its p99 latency is at most 0.73 ms, below even the 1 ms interval, and at most 0.24\% of arrivals are late. The maximum dropped share over pairs is 0.33\% at 1 ms and 0.06--0.11\% at 1.5--3 ms. These drops come from a few pauses of the replay process, not from RESYNC's service time. The longest stall at each interval lasts 75--123 ms, and a replay's drops often come from one of its three repetitions. The compact models are hit by the same kind of pause, with stalls of up to 66 ms. A pause of fixed duration drops more arrivals at shorter intervals. For the FI-2010 InceptionTime, for example, the longest stall covers 93, 60, 41 and 31 consecutive arrivals at 1, 1.5, 2 and 2.5 ms, 78--93 ms each time. Because drops depend on when such pauses happen, the drop rate of a single pair need not decrease with $\Delta$.

\paragraph{Anchor ages across intervals.} Anchor ages scale with $1/\Delta$. The per-pair median falls from 7--15 observations at 1 ms to 3--4 at 3 ms, and the largest p99 from 20 to 5. At no interval do more than 0.004\% of any pair's predictions use $n=1$. The share with $n\leq2$ is at most 0.001\% at 1 and 1.5 ms and at most 4.7\% at 2 ms. At 2.5 ms it is about half in five pairs, and at 3 ms about half in eleven pairs, all except the three Mamba-2 teachers and the FI-2010 TSMixer.

\paragraph{Clamping.} Clamped predictions occur only in the tail after a stall. There are 98 at 1 ms (seven pairs, at most 0.066\% of a pair's predictions, largest age 80), 26 at 1.5 ms (at most 0.012\%), 5 at 2 ms (at most 0.003\%) and none at 2.5 and 3 ms. 126 of these 129 predictions lie within 40 arrivals of a dropped arrival: the pause also delays the completion of the next anchor, and the age returns to its usual range once that anchor lands. Their quality is not covered by the offline evaluation, but at these shares they cannot change the quality of the delivered predictions measurably.

\begin{table}[h]
\caption{Live replay per pair at $\Delta=1$, 1.5, 2, 2.5 and 3 ms. Teacher coverage: fraction of arrivals that received their own prediction. Compact and RESYNC miss: share of arrivals without a prediction within one interval. RESYNC dropped: share of arrivals already one interval late when served. p99: RESYNC's 99th-percentile latency of completed predictions. Anchor age: median [5th--95th percentile], maximum, and share of predictions with $n\leq2$. $n>n_{\max}$: number of clamped RESYNC predictions (of about 74{,}000).}
\label{tab:intervals}
\centering
\resizebox{\textwidth}{!}{%
\begin{tabular}{llcccccccccc}
\toprule
 & & & Teacher & Compact & \multicolumn{3}{c}{RESYNC service} & \multicolumn{4}{c}{RESYNC anchor age} \\
\cmidrule(lr){4-4}\cmidrule(lr){5-5}\cmidrule(lr){6-8}\cmidrule(lr){9-12}
Dataset & Architecture & $\Delta$ (ms) & Coverage & Miss (\%) & Dropped (\%) & Miss (\%) & p99 (ms) & Median [p5--p95] & Max & $n\leq2$ (\%) & $n>n_{\max}$ \\
\midrule
FI-2010 & Transformer & 1 & 0.24 & 0.22 & 0.19 & 0.27 & 0.61 & 7 [5--9] & 73 & 0.00 & 9 \\
 &  & 1.5 & 0.36 & 0.07 & 0.06 & 0.11 & 0.59 & 4 [3--5] & 13 & 0.00 & 0 \\
 &  & 2 & 0.49 & 0.11 & 0.02 & 0.04 & 0.57 & 4 [3--5] & 10 & 0.33 & 0 \\
 &  & 2.5 & 0.47 & 0.00 & 0.04 & 0.04 & 0.67 & 3 [2--5] & 31 & 33.14 & 0 \\
 &  & 3 & 0.62 & 0.00 & 0.03 & 0.03 & 0.60 & 3 [2--3] & 27 & 49.92 & 0 \\
 & DeepConvLSTM & 1 & 0.19 & 0.01 & 0.04 & 0.14 & 0.55 & 9 [6--11] & 19 & 0.00 & 0 \\
 &  & 1.5 & 0.28 & 0.01 & 0.01 & 0.01 & 0.57 & 6 [4--7] & 13 & 0.00 & 0 \\
 &  & 2 & 0.38 & 0.02 & 0.00 & 0.01 & 0.56 & 4 [3--5] & 7 & 0.00 & 0 \\
 &  & 2.5 & 0.48 & 0.01 & 0.00 & 0.01 & 0.55 & 4 [3--5] & 6 & 0.06 & 0 \\
 &  & 3 & 0.56 & 0.03 & 0.00 & 0.00 & 0.64 & 3 [2--3] & 5 & 49.95 & 0 \\
 & InceptionTime & 1 & 0.24 & 0.00 & 0.13 & 0.13 & 0.67 & 7 [5--9] & 37 & 0.00 & 3 \\
 &  & 1.5 & 0.38 & 0.00 & 0.08 & 0.09 & 0.69 & 4 [3--6] & 27 & 0.00 & 0 \\
 &  & 2 & 0.52 & 0.00 & 0.06 & 0.06 & 0.66 & 4 [3--5] & 24 & 4.74 & 0 \\
 &  & 2.5 & 0.65 & 0.00 & 0.04 & 0.04 & 0.64 & 3 [2--3] & 18 & 50.00 & 0 \\
 &  & 3 & 0.71 & 0.03 & 0.04 & 0.06 & 0.65 & 3 [2--3] & 16 & 49.93 & 0 \\
 & Mamba-2 & 1 & 0.14 & 100.00 & 0.08 & 0.14 & 0.69 & 14 [9--18] & 41 & 0.00 & 11 \\
 &  & 1.5 & 0.24 & 0.06 & 0.06 & 0.09 & 0.66 & 7 [5--10] & 30 & 0.00 & 0 \\
 &  & 2 & 0.32 & 0.04 & 0.00 & 0.01 & 0.71 & 6 [4--7] & 14 & 0.00 & 0 \\
 &  & 2.5 & 0.43 & 0.00 & 0.00 & 0.00 & 0.72 & 4 [3--5] & 6 & 0.00 & 0 \\
 &  & 3 & 0.46 & 0.01 & 0.03 & 0.05 & 0.68 & 4 [3--5] & 21 & 0.05 & 0 \\
 & TSMixer & 1 & 0.14 & 0.32 & 0.02 & 0.08 & 0.71 & 15 [10--19] & 31 & 0.00 & 0 \\
 &  & 1.5 & 0.20 & 0.08 & 0.11 & 0.14 & 0.72 & 9 [6--12] & 46 & 0.00 & 9 \\
 &  & 2 & 0.29 & 0.00 & 0.05 & 0.06 & 0.66 & 6 [4--8] & 41 & 0.00 & 2 \\
 &  & 2.5 & 0.34 & 0.03 & 0.00 & 0.00 & 0.69 & 5 [4--7] & 7 & 0.00 & 0 \\
 &  & 3 & 0.43 & 0.03 & 0.00 & 0.00 & 0.66 & 4 [3--5] & 5 & 0.00 & 0 \\
\midrule
Synthetic-1 & Transformer & 1 & 0.25 & 0.00 & 0.11 & 0.11 & 0.58 & 7 [5--9] & 80 & 0.00 & 10 \\
 &  & 1.5 & 0.37 & 0.00 & 0.07 & 0.07 & 0.56 & 4 [3--5] & 43 & 0.00 & 4 \\
 &  & 2 & 0.51 & 0.04 & 0.00 & 0.00 & 0.65 & 4 [3--5] & 5 & 1.40 & 0 \\
 &  & 2.5 & 0.54 & 0.04 & 0.07 & 0.09 & 0.57 & 3 [2--3] & 29 & 49.58 & 0 \\
 &  & 3 & 0.67 & 0.04 & 0.03 & 0.05 & 0.58 & 3 [2--3] & 24 & 49.93 & 0 \\
 & DeepConvLSTM & 1 & 0.19 & 0.04 & 0.06 & 0.11 & 0.56 & 8 [6--11] & 24 & 0.00 & 0 \\
 &  & 1.5 & 0.29 & 0.08 & 0.01 & 0.04 & 0.53 & 6 [4--7] & 13 & 0.00 & 0 \\
 &  & 2 & 0.36 & 0.00 & 0.00 & 0.00 & 0.61 & 4 [3--5] & 5 & 0.00 & 0 \\
 &  & 2.5 & 0.44 & 0.00 & 0.00 & 0.00 & 0.62 & 4 [3--5] & 5 & 0.00 & 0 \\
 &  & 3 & 0.54 & 0.01 & 0.00 & 0.01 & 0.59 & 3 [2--3] & 6 & 49.94 & 0 \\
 & InceptionTime & 1 & 0.19 & 0.33 & 0.17 & 0.19 & 0.71 & 9 [6--11] & 25 & 0.00 & 0 \\
 &  & 1.5 & 0.38 & 0.13 & 0.08 & 0.11 & 0.67 & 4 [3--6] & 43 & 0.00 & 5 \\
 &  & 2 & 0.49 & 0.02 & 0.05 & 0.07 & 0.66 & 4 [3--5] & 29 & 2.28 & 0 \\
 &  & 2.5 & 0.62 & 0.05 & 0.00 & 0.01 & 0.63 & 3 [2--3] & 7 & 49.92 & 0 \\
 &  & 3 & 0.64 & 0.03 & 0.00 & 0.00 & 0.71 & 3 [2--3] & 16 & 49.60 & 0 \\
 & Mamba-2 & 1 & 0.14 & 99.82 & 0.18 & 0.37 & 0.68 & 13 [9--19] & 57 & 0.00 & 49 \\
 &  & 1.5 & 0.24 & 0.08 & 0.01 & 0.01 & 0.71 & 8 [5--10] & 19 & 0.00 & 0 \\
 &  & 2 & 0.34 & 0.01 & 0.00 & 0.00 & 0.70 & 6 [4--7] & 9 & 0.00 & 0 \\
 &  & 2.5 & 0.39 & 0.01 & 0.00 & 0.01 & 0.68 & 4 [3--5] & 10 & 0.00 & 0 \\
 &  & 3 & 0.49 & 0.00 & 0.00 & 0.00 & 0.67 & 4 [3--5] & 5 & 0.06 & 0 \\
 & TSMixer & 1 & 0.22 & 0.00 & 0.12 & 0.12 & 0.65 & 9 [6--13] & 44 & 0.00 & 5 \\
 &  & 1.5 & 0.33 & 0.04 & 0.09 & 0.11 & 0.65 & 6 [4--7] & 44 & 0.00 & 3 \\
 &  & 2 & 0.35 & 0.03 & 0.02 & 0.03 & 0.73 & 4 [3--6] & 22 & 0.00 & 0 \\
 &  & 2.5 & 0.51 & 0.08 & 0.00 & 0.01 & 0.68 & 4 [3--5] & 10 & 1.07 & 0 \\
 &  & 3 & 0.60 & 0.04 & 0.00 & 0.01 & 0.69 & 3 [2--3] & 7 & 49.84 & 0 \\
\midrule
Synthetic-2 & Transformer & 1 & 0.24 & 0.10 & 0.01 & 0.01 & 0.57 & 7 [5--9] & 12 & 0.00 & 0 \\
 &  & 1.5 & 0.36 & 0.02 & 0.08 & 0.09 & 0.66 & 4 [3--6] & 50 & 0.00 & 2 \\
 &  & 2 & 0.48 & 0.00 & 0.07 & 0.08 & 0.58 & 4 [3--5] & 36 & 0.00 & 2 \\
 &  & 2.5 & 0.63 & 0.00 & 0.04 & 0.04 & 0.56 & 3 [2--3] & 31 & 49.96 & 0 \\
 &  & 3 & 0.69 & 0.02 & 0.06 & 0.10 & 0.60 & 3 [2--3] & 24 & 49.82 & 0 \\
 & DeepConvLSTM & 1 & 0.19 & 0.21 & 0.06 & 0.11 & 0.56 & 9 [6--11] & 22 & 0.00 & 0 \\
 &  & 1.5 & 0.29 & 0.00 & 0.00 & 0.00 & 0.58 & 6 [4--7] & 9 & 0.00 & 0 \\
 &  & 2 & 0.39 & 0.07 & 0.02 & 0.05 & 0.55 & 4 [3--5] & 11 & 0.00 & 0 \\
 &  & 2.5 & 0.48 & 0.01 & 0.00 & 0.01 & 0.57 & 4 [3--5] & 8 & 0.00 & 0 \\
 &  & 3 & 0.58 & 0.02 & 0.01 & 0.01 & 0.56 & 3 [2--3] & 7 & 49.94 & 0 \\
 & InceptionTime & 1 & 0.23 & 0.79 & 0.07 & 0.14 & 0.65 & 7 [5--9] & 21 & 0.00 & 0 \\
 &  & 1.5 & 0.39 & 0.16 & 0.03 & 0.08 & 0.63 & 4 [3--5] & 12 & 0.00 & 0 \\
 &  & 2 & 0.42 & 0.00 & 0.06 & 0.07 & 0.65 & 4 [3--5] & 17 & 3.62 & 0 \\
 &  & 2.5 & 0.63 & 0.07 & 0.04 & 0.04 & 0.65 & 3 [2--3] & 16 & 47.38 & 0 \\
 &  & 3 & 0.78 & 0.01 & 0.03 & 0.04 & 0.63 & 3 [2--3] & 22 & 49.99 & 0 \\
 & Mamba-2 & 1 & 0.17 & 91.73 & 0.33 & 0.57 & 0.69 & 13 [9--17] & 21 & 0.00 & 0 \\
 &  & 1.5 & 0.22 & 2.71 & 0.00 & 0.01 & 0.70 & 8 [6--11] & 23 & 0.00 & 0 \\
 &  & 2 & 0.34 & 0.02 & 0.01 & 0.03 & 0.66 & 6 [4--7] & 15 & 0.00 & 0 \\
 &  & 2.5 & 0.38 & 0.00 & 0.00 & 0.00 & 0.70 & 4 [3--6] & 8 & 0.00 & 0 \\
 &  & 3 & 0.50 & 0.00 & 0.02 & 0.03 & 0.71 & 4 [3--5] & 28 & 0.00 & 0 \\
 & TSMixer & 1 & 0.22 & 0.00 & 0.27 & 0.37 & 0.67 & 9 [6--13] & 42 & 0.00 & 11 \\
 &  & 1.5 & 0.29 & 0.06 & 0.00 & 0.00 & 0.68 & 6 [4--8] & 34 & 0.00 & 3 \\
 &  & 2 & 0.47 & 0.00 & 0.06 & 0.06 & 0.66 & 4 [3--5] & 33 & 0.00 & 1 \\
 &  & 2.5 & 0.48 & 0.00 & 0.04 & 0.04 & 0.69 & 3 [2--5] & 26 & 19.46 & 0 \\
 &  & 3 & 0.60 & 0.02 & 0.05 & 0.06 & 0.66 & 3 [2--3] & 21 & 49.90 & 0 \\
\bottomrule
\end{tabular}
}
\end{table}

\section{Runtime and resource cost}
\label{app:cost}
\paragraph{Controlled sweep.} For every pair we time the complete per-arrival tick of the live replay---two encoder steps (the active and the candidate state), synchronization and the head---at fixed anchor gaps $n\in\{1,2,3,4,6,8,12,16,20,24,28,32,40,48,64,80\}$, on one CPU thread with garbage collection disabled and nothing else running. Each gap is timed with 2{,}000 calls in each of three sessions. Above $n_{\max}=32$ the updater uses the gap embedding of $n_{\max}$, as in the live replay. For comparison we time the same update when the GRU re-encodes the whole suffix at every arrival. Before timing, the incremental tick is checked against re-encoding at every tested gap. Table~\ref{tab:runtime} reports the median over the sessions and a least-squares slope against $n$, together with the anchor switches of the live replay at $\Delta=1$ ms.

\paragraph{Resource replay.} We repeat the replay at $\Delta=1$ ms with each method and repetition in its own process, three repetitions each, and record CPU time by thread, peak host memory (reset just before the replay), PyTorch's peak GPU allocation, and the fraction of the replay during which a GPU call is in flight between CUDA events (GPU busy). Table~\ref{tab:resources} reports the median over repetitions and, for memory, the worst repetition. The compact model's process never initializes CUDA, and the RESYNC process holds one copy of the teacher on the GPU.

\paragraph{Per-arrival cost.} The complete RESYNC tick takes 0.22--0.37 ms on one CPU thread and is flat from $n=1$ to $n=80$, beyond $n_{\max}$ included: its median changes by less than 0.01~$\mu$s per observation of anchor age (Table~\ref{tab:runtime}). Re-encoding the observed suffix with the GRU at every arrival instead grows by 23--27~$\mu$s per observation, so it is 3.5--4.6$\times$ slower than the incremental tick at $n=32$ and 6.8--9.6$\times$ slower at $n=80$. Computing the anchor-dependent term once per anchor saves 0.02--0.06 ms per tick.

\paragraph{Anchor switches.} In live replay at 1 ms the body is relaunched every 5--10 arrivals (99--200 launches per 1{,}000 arrivals), so 10--20\% of ticks also install a new anchor. These anchor-switch ticks have a median latency of 0.48--0.61 ms, against 0.35--0.62 ms for the other ticks, and a p99 of at most 0.74 ms against 0.71 ms. They miss the 1 ms deadline no more often: at most 0.20\% of them, against 0.63\% of the other ticks.

\paragraph{Resources.} Replayed at 1 ms with each method in its own process (Table~\ref{tab:resources}), the RESYNC system uses 1.21--1.37 CPU cores, of which 0.87--0.93 are the body worker thread waiting on the GPU, against 1.00 for the synchronous teacher and 0.40--0.99 for the compact model. Because the worker relaunches the body as soon as it finishes, RESYNC keeps the GPU busy for 85--93\% of the replay, close to the synchronous teacher's 98--99\%. Its peak host memory is within 0.02 GB of the teacher's (1.06--1.34 GB, against 0.65--0.70 GB for the compact model), and its peak GPU allocation exceeds the teacher's by at most 33 MB. The updater holds 0.20--0.68M parameters (1.5--5.2 MB with its precomputed cache), and each anchor adds 3--5 KB of state. The instrumented replays serve arrivals as in Table~\ref{tab:live}, with RESYNC coverage of at least 0.997 and median latency within 0.07 ms, except that the Mamba-2 compact models run faster in a process of their own (median 1.03--1.04 ms against 1.24--1.60 ms), though still above the 1 ms interval.

\begin{table}[h]
\caption{Runtime of the RESYNC tick. Controlled: median tick time over all tested gaps (it varies by less than 0.01 ms), least-squares slope of the tick and of re-encoding the suffix against $n$ in $\mu$s per observation, and re-encoding over the tick at $n=32$ and $n=80$. Live replay at $\Delta=1$ ms: body launches per 1{,}000 arrivals, the share of ticks that install a new anchor, and the p99 latency in ms of the other ticks and of these anchor-switch ticks.}
\label{tab:runtime}
\centering
\resizebox{\textwidth}{!}{%
\begin{tabular}{llccccccccc}
\toprule
 & & \multicolumn{5}{c}{Controlled (s3), all gaps $n\leq80$} & \multicolumn{4}{c}{Live replay, $\Delta=1$ ms} \\
\cmidrule(lr){3-7}\cmidrule(lr){8-11}
Dataset & Architecture & Tick p50 (ms) & Tick slope & Re-enc.\ slope & Re-enc./tick, $n=32$ & $n=80$ & Launches / 1k & Switch ticks (\%) & p99 other & p99 switch \\
\midrule
FI-2010 & Transformer & 0.37 & -0.002 & 26.9 & 3.6 & 6.8 & 199 & 20 & 0.53 & 0.65 \\
 & DeepConvLSTM & 0.33 & +0.005 & 26.8 & 3.7 & 7.3 & 166 & 17 & 0.42 & 0.58 \\
 & InceptionTime & 0.25 & +0.004 & 23.9 & 4.2 & 8.4 & 199 & 20 & 0.66 & 0.67 \\
 & Mamba-2 & 0.22 & +0.000 & 24.3 & 4.6 & 9.6 & 107 & 11 & 0.69 & 0.71 \\
 & TSMixer & 0.27 & -0.003 & 23.8 & 3.9 & 7.9 & 99 & 10 & 0.71 & 0.70 \\
\midrule
Synthetic-1 & Transformer & 0.25 & +0.002 & 23.7 & 4.1 & 8.3 & 200 & 20 & 0.53 & 0.66 \\
 & DeepConvLSTM & 0.28 & +0.000 & 24.7 & 3.9 & 7.9 & 168 & 17 & 0.51 & 0.66 \\
 & InceptionTime & 0.24 & -0.003 & 24.2 & 4.3 & 8.9 & 166 & 17 & 0.70 & 0.74 \\
 & Mamba-2 & 0.23 & +0.001 & 24.3 & 4.5 & 9.4 & 110 & 11 & 0.68 & 0.70 \\
 & TSMixer & 0.31 & +0.003 & 25.2 & 3.7 & 7.3 & 150 & 15 & 0.65 & 0.67 \\
\midrule
Synthetic-2 & Transformer & 0.35 & +0.009 & 25.3 & 3.5 & 6.8 & 200 & 20 & 0.54 & 0.58 \\
 & DeepConvLSTM & 0.26 & +0.001 & 23.4 & 4.0 & 8.0 & 167 & 17 & 0.47 & 0.58 \\
 & InceptionTime & 0.24 & +0.005 & 24.4 & 4.4 & 9.0 & 197 & 20 & 0.63 & 0.68 \\
 & Mamba-2 & 0.22 & +0.007 & 23.9 & 4.5 & 9.3 & 110 & 11 & 0.69 & 0.69 \\
 & TSMixer & 0.25 & +0.002 & 23.7 & 4.1 & 8.4 & 152 & 15 & 0.66 & 0.68 \\
\bottomrule
\end{tabular}
}
\end{table}

\begin{table}[h]
\caption{Resources at $\Delta=1$ ms, each method in its own process. CPU cores: process CPU time over wall time, with the RESYNC body worker thread in parentheses. GPU busy: share of the replay with a GPU call in flight. Peak host memory: resident set size. Peak GPU memory: PyTorch's peak allocation, CUDA context excluded.}
\label{tab:resources}
\centering
\resizebox{\textwidth}{!}{%
\begin{tabular}{llcccccccccc}
\toprule
 & & \multicolumn{3}{c}{CPU cores} & \multicolumn{2}{c}{GPU busy (\%)} & \multicolumn{3}{c}{Peak host memory (GB)} & \multicolumn{2}{c}{Peak GPU memory (MB)} \\
\cmidrule(lr){3-5}\cmidrule(lr){6-7}\cmidrule(lr){8-10}\cmidrule(lr){11-12}
Dataset & Architecture & Teacher & Compact & RESYNC (worker) & Teacher & RESYNC & Teacher & Compact & RESYNC & Teacher & RESYNC \\
\midrule
FI-2010 & Transformer & 1.00 & 0.64 & 1.26 (0.89) & 98 & 87 & 1.07 & 0.67 & 1.08 & 66 & 99 \\
 & DeepConvLSTM & 1.00 & 0.40 & 1.21 (0.87) & 98 & 85 & 1.19 & 0.66 & 1.21 & 172 & 172 \\
 & InceptionTime & 1.00 & 0.63 & 1.37 (0.92) & 98 & 92 & 1.23 & 0.66 & 1.24 & 45 & 45 \\
 & Mamba-2 & 1.00 & 0.99 & 1.37 (0.87) & 98 & 91 & 1.32 & 0.67 & 1.33 & 36 & 69 \\
 & TSMixer & 1.00 & 0.82 & 1.35 (0.89) & 99 & 93 & 1.09 & 0.66 & 1.10 & 157 & 190 \\
\midrule
Synthetic-1 & Transformer & 1.00 & 0.65 & 1.29 (0.92) & 98 & 90 & 1.04 & 0.65 & 1.06 & 66 & 99 \\
 & DeepConvLSTM & 1.00 & 0.42 & 1.22 (0.87) & 98 & 86 & 1.15 & 0.65 & 1.17 & 172 & 172 \\
 & InceptionTime & 1.00 & 0.61 & 1.32 (0.88) & 98 & 87 & 1.22 & 0.65 & 1.23 & 45 & 45 \\
 & Mamba-2 & 1.00 & 0.99 & 1.36 (0.89) & 98 & 92 & 1.29 & 0.65 & 1.30 & 36 & 69 \\
 & TSMixer & 1.00 & 0.44 & 1.35 (0.89) & 98 & 90 & 1.08 & 0.65 & 1.09 & 59 & 92 \\
\midrule
Synthetic-2 & Transformer & 1.00 & 0.73 & 1.30 (0.93) & 98 & 91 & 1.08 & 0.70 & 1.10 & 66 & 99 \\
 & DeepConvLSTM & 1.00 & 0.41 & 1.22 (0.87) & 98 & 85 & 1.19 & 0.69 & 1.20 & 172 & 172 \\
 & InceptionTime & 1.00 & 0.63 & 1.37 (0.93) & 98 & 92 & 1.25 & 0.69 & 1.26 & 45 & 45 \\
 & Mamba-2 & 1.00 & 0.98 & 1.34 (0.88) & 98 & 91 & 1.33 & 0.70 & 1.34 & 36 & 69 \\
 & TSMixer & 1.00 & 0.45 & 1.35 (0.88) & 98 & 89 & 1.12 & 0.69 & 1.14 & 59 & 92 \\
\bottomrule
\end{tabular}
}
\end{table}

\section{Recursive latent-delta study}
\label{app:recursive}
\paragraph{Protocol.} The recursive updater is the RESYNC module with the same capacity, objective, training budget and label-free selection, trained only on exact one-step pairs $(z_{t-1},x_t)\mapsto z_t$, as Delta Distillation trains students on the teacher's one-step deltas~\citep{habibian-2022-delta}. At inference it chains $n$ updates from the exact anchor, each applied to its own previous estimate. It is not a reproduction of Delta Distillation's blockwise architecture. Both updaters are trained afresh for this study and scored on 10{,}000 fixed test targets at every gap from 1 to 32, with the macro-F1, the relative latent error $\|\widehat z-z_t\|/\|z_t\|$ and the agreement with the fresh teacher.

\begin{table}[h]
\caption{RESYNC (R) versus the matched recursive latent-delta updater (Rec.) on 10{,}000 fixed test targets: macro-F1 at anchor gaps $n=1$, 8 and 32, and relative latent error $\|\widehat z-z_t\|/\|z_t\|$ at $n=32$. Fresh is the teacher's F1 on the same targets.}
\label{tab:recursive}
\centering
\resizebox{\textwidth}{!}{%
\begin{tabular}{llccccccccc}
\toprule
 & & & \multicolumn{2}{c}{$n=1$} & \multicolumn{2}{c}{$n=8$} & \multicolumn{2}{c}{$n=32$} & \multicolumn{2}{c}{Latent error, $n=32$} \\
\cmidrule(lr){4-5}\cmidrule(lr){6-7}\cmidrule(lr){8-9}\cmidrule(lr){10-11}
Dataset & Architecture & Fresh & R & Rec. & R & Rec. & R & Rec. & R & Rec. \\
\midrule
FI-2010 & Transformer & 0.699 & 0.531 & 0.657 & 0.707 & 0.607 & 0.706 & 0.486 & 0.27 & 0.65 \\
 & DeepConvLSTM & 0.664 & 0.495 & 0.506 & 0.685 & 0.422 & 0.680 & 0.306 & 0.24 & 2.32 \\
 & InceptionTime & 0.660 & 0.479 & 0.482 & 0.670 & 0.431 & 0.669 & 0.402 & 0.16 & 0.69 \\
 & Mamba-2 & 0.684 & 0.560 & 0.660 & 0.695 & 0.623 & 0.693 & 0.599 & 0.42 & 0.70 \\
 & TSMixer & 0.687 & 0.499 & 0.589 & 0.698 & 0.550 & 0.695 & 0.558 & 0.32 & 0.55 \\
\midrule
Synthetic-1 & Transformer & 0.937 & 0.916 & 0.911 & 0.859 & 0.464 & 0.795 & 0.368 & 0.58 & 1.20 \\
 & DeepConvLSTM & 0.931 & 0.924 & 0.922 & 0.923 & 0.722 & 0.906 & 0.323 & 0.38 & 4.64 \\
 & InceptionTime & 0.940 & 0.929 & 0.924 & 0.892 & 0.462 & 0.855 & 0.322 & 0.20 & 5.20 \\
 & Mamba-2 & 0.941 & 0.932 & 0.940 & 0.922 & 0.880 & 0.927 & 0.637 & 0.33 & 0.71 \\
 & TSMixer & 0.941 & 0.934 & 0.942 & 0.904 & 0.852 & 0.897 & 0.582 & 0.37 & 0.58 \\
\midrule
Synthetic-2 & Transformer & 0.778 & 0.747 & 0.767 & 0.660 & 0.617 & 0.615 & 0.452 & 0.41 & 0.63 \\
 & DeepConvLSTM & 0.606 & 0.597 & 0.601 & 0.605 & 0.586 & 0.609 & 0.458 & 0.11 & 2.76 \\
 & InceptionTime & 0.650 & 0.637 & 0.649 & 0.633 & 0.585 & 0.608 & 0.488 & 0.25 & 0.54 \\
 & Mamba-2 & 0.596 & 0.594 & 0.596 & 0.594 & 0.594 & 0.593 & 0.593 & 0.13 & 0.20 \\
 & TSMixer & 0.598 & 0.596 & 0.597 & 0.596 & 0.592 & 0.599 & 0.590 & 0.19 & 0.26 \\
\bottomrule
\end{tabular}

}
\end{table}

\paragraph{Results.} At $n=1$ the recursive updater is better in twelve of the fifteen pairs, and at $n=2$ and 3 RESYNC is better in nine. From $n=4$ onwards RESYNC is better in 13--15 pairs at every gap. Where the recursive updater is still ahead from $n=4$, the two differ by at most 0.003: at several gaps of Synthetic-2 Mamba-2, where both stay within 0.004 of the fresh teacher throughout, and at single gaps of the Synthetic-1 and Synthetic-2 TSMixer. At $n=32$, accumulated recursive error is larger than simply not updating in nine of the fifteen pairs. In the worst case, the Synthetic-1 InceptionTime, it reaches 5.20 against 0.39 for the stale anchor. There, and for the FI-2010 and Synthetic-1 DeepConvLSTM, the recursive updater's macro-F1 at $n=32$ falls below the stale teacher's. Validation shows the same difference under the label-free selection criterion: the selected RESYNC checkpoints remove 0.77--0.98 of the stale validation loss, the recursive ones between $-3.37$ and 0.95.

\section{Incremental-inference alternatives}
\label{app:streaming}
\paragraph{Models.} Both families are trained with the teachers' recipe and evaluated at every test position.
\begin{itemize}
    \item \textbf{Continual Transformer}~\citep{hedegaard-2023-continual}: a pre-norm encoder with non-causal attention and last-step readout. With one block it uses Continual Single-Output attention over cached keys and values. With two blocks, the first uses Continual Retroactive attention, which adds the newest key and removes the oldest from each token's cached softmax numerator and denominator, and the second uses Single-Output attention. We use rotary position embeddings instead of the original Recycling Positional Encoding, so attention depends only on relative position and streaming is exact without random training-time offsets. Deeper stacks have no exact streaming form, so the sizes grow in width: $d=8$ to 512, with one block for mini and small and two otherwise.
    \item \textbf{Mamba-2, recurrent}: the windowed Mamba-2 run as a recurrence. Its state starts on the first window of each contiguous test segment and is never reset. It therefore only approximates the windowed model, which restarts from a zero state at every window.
\end{itemize}

\paragraph{Choosing the sizes in Table~\ref{tab:incremental}.} Each incremental model is timed in three ways: one step on one CPU thread, as RESYNC's foreground path; one step on the GPU; and re-encoding the full window on the GPU. Each size is represented by the fastest of the three, which favours the incremental models over RESYNC. For each family we then take the largest size whose fastest time per arrival is no slower than RESYNC's median time on the same teacher family, and the next size up. The choice is made on speed alone, before looking at F1. The CPU step is the fastest option for every model in Table~\ref{tab:incremental} except the next-size Continual Transformer, which is fastest when re-encoding its window on the GPU; the GPU is never fast enough at RESYNC's speed, since even the smallest models take 0.55--0.97 ms per arrival there. The size grid has no variants with times between about 0.33 and 0.90 ms, so the first model sits somewhat below RESYNC's speed and the second well above it.

\paragraph{Comparability.} All timings use the replay protocol of Section~\ref{sec:experimental-design} on the same type of node. RESYNC was timed with the teacher body running concurrently on the GPU, which if anything slows RESYNC. The incremental models are scored at every test position, while RESYNC is scored on the fixed targets of Appendix~\ref{app:quality-details}, which cover all but the first 32 positions of each test day or run. The incremental models were replayed at $\Delta=3$ ms and RESYNC at $\Delta=1$ ms; the service time of a model that keeps up hardly depends on the interval (RESYNC's median latency is 0.35--0.62 ms at 1 ms and 0.43--0.59 ms at 3 ms).

\begin{table}[h]
\caption{Incremental inference at RESYNC's speed. Each family is compared with RESYNC on the teacher of the same family. For the incremental models we show the largest size no slower than RESYNC and the next size up. Time is the median service time per arrival. RESYNC runs on one CPU thread; each incremental model is timed as one step on one CPU thread, one step on the GPU and re-encoding the full window on the GPU, and shown at its fastest (the CPU step unless marked); the synchronous teachers run on the GPU. RESYNC's macro-F1 is at the tested anchor gap nearest the median age it sees in live replay at $\Delta=1$ ms ($n=8$ for the Transformer; for Mamba-2, $n=16$ on FI-2010 and $n=12$ on the synthetic streams).}
\label{tab:incremental}
\centering
\small
\begin{tabular}{llcc}
\toprule
Method & Size & Time (ms) & Macro-F1 \\
\midrule
\multicolumn{4}{l}{\emph{FI-2010}} \\
RESYNC on the Transformer teacher & -- & 0.38 & 0.688 \\
Continual Transformer (exact) & small & 0.27 & 0.603 \\
Continual Transformer (exact) & tiny & 1.22 (GPU, window) & 0.654 \\
Synchronous Transformer teacher & large & 4.06 (GPU) & 0.676 \\
RESYNC on the Mamba-2 teacher & -- & 0.51 & 0.683 \\
Mamba-2, recurrent (approximate) & small & 0.29 & 0.646 \\
Mamba-2, recurrent (approximate) & tiny & 1.00 & 0.677 \\
Synchronous Mamba-2 teacher & base & 6.89 (GPU) & 0.670 \\
\midrule
\multicolumn{4}{l}{\emph{Synthetic-1}} \\
RESYNC on the Transformer teacher & -- & 0.42 & 0.862 \\
Continual Transformer (exact) & small & 0.29 & 0.462 \\
Continual Transformer (exact) & tiny & 1.33 (GPU, window) & 0.919 \\
Synchronous Transformer teacher & large & 3.95 (GPU) & 0.937 \\
RESYNC on the Mamba-2 teacher & -- & 0.49 & 0.927 \\
Mamba-2, recurrent (approximate) & small & 0.32 & 0.540 \\
Mamba-2, recurrent (approximate) & tiny & 0.97 & 0.933 \\
Synchronous Mamba-2 teacher & base & 6.50 (GPU) & 0.940 \\
\midrule
\multicolumn{4}{l}{\emph{Synthetic-2}} \\
RESYNC on the Transformer teacher & -- & 0.42 & 0.667 \\
Continual Transformer (exact) & small & 0.30 & 0.583 \\
Continual Transformer (exact) & tiny & 1.03 (GPU, window) & 0.621 \\
Synchronous Transformer teacher & large & 4.11 (GPU) & 0.774 \\
RESYNC on the Mamba-2 teacher & -- & 0.50 & 0.590 \\
Mamba-2, recurrent (approximate) & small & 0.28 & 0.588 \\
Mamba-2, recurrent (approximate) & tiny & 0.90 & 0.597 \\
Synchronous Mamba-2 teacher & base & 5.92 (GPU) & 0.588 \\
\bottomrule
\end{tabular}

\end{table}

\paragraph{At RESYNC's speed.} The largest incremental models no slower than RESYNC take 0.27--0.32 ms per arrival, against RESYNC's 0.38--0.51 ms (Table~\ref{tab:incremental}). They trail RESYNC by 0.085 (Continual Transformer) and 0.037 (recurrent Mamba-2) macro-F1 on FI-2010, by 0.39--0.40 on Synthetic-1 and by 0.084 for the Continual Transformer on Synthetic-2. Only recurrent Mamba-2 on Synthetic-2, whose teacher has no advantage over the fast models, comes within 0.002.

\paragraph{One size up.} The incremental models then take 0.90--1.33 ms per arrival at their fastest, 1.8--3.2$\times$ RESYNC's time per arrival; the Continual Transformer is fastest when it re-encodes its window on the GPU, the recurrent Mamba-2 when it steps on the CPU. On FI-2010, recurrent Mamba-2 reaches 0.677 against RESYNC's 0.683, and the Continual Transformer 0.654 against 0.688. On Synthetic-1 both pass RESYNC: the Continual Transformer reaches 0.919 against 0.862, and recurrent Mamba-2 0.933 against 0.927. On Synthetic-2, recurrent Mamba-2 reaches 0.597 against 0.590, while the Continual Transformer stays at 0.621, below RESYNC's 0.667 and the teacher's 0.774. These steps do not fit the 1 ms interval of Table~\ref{tab:live}: a Continual Transformer taking 1.03--1.33 ms per arrival cannot keep up, and a 0.90--1.00 ms recurrent Mamba-2 step leaves almost no headroom. At equal speed, RESYNC's advantage grows where quality depends on capacity, from 0.04--0.09 macro-F1 on FI-2010 to about 0.4 on Synthetic-1. The step between capacity levels is steep, however: on Synthetic-1 an 18k-parameter Continual Transformer and a 32k-parameter recurrent Mamba-2 already reach 0.92--0.93. RESYNC's small excess over its FI-2010 teachers is the class-frequency effect of Appendix~\ref{app:above-teacher}; measured against the teachers instead, the conclusion is the same.

\paragraph{GPU and larger sizes.} On the GPU, the Continual Transformer's incremental step takes 1.6--2.3 ms from the tiny size upwards, and re-encoding its full window 1.0--1.3 ms; neither grows much with size (Table~\ref{tab:streaming}). At $\Delta=3$ ms every size therefore keeps up except the Synthetic-1 xlarge model, which drops 7\% of arrivals, and the largest sizes reach 0.671, 0.938 and 0.750 macro-F1 on FI-2010, Synthetic-1 and Synthetic-2. Neither GPU option would keep up with the 1 ms arrivals of Table~\ref{tab:live}. If a GPU is available for the per-arrival path and the stream is slow enough, a shallow exact-streaming Transformer is therefore a viable alternative. Recurrent Mamba-2 keeps up at 3 ms up to the base size on the CPU and up to the tiny size on the GPU. A stream model cannot skip observations: when a step takes longer than $\Delta$, its backlog grows without bound and every arrival counts as dropped.

\begin{table}[h]
\caption{Full size grid on every dataset: windowed and streamed test macro-F1, agreement between them, relative latent error of the streamed latent, and median service time in ms per arrival with the fraction of dropped arrivals at $\Delta=3$ ms in parentheses. ``Step'' is one incremental update; ``Window, GPU'' re-encodes the full window.}
\label{tab:streaming}
\centering
\resizebox{\textwidth}{!}{%
\begin{tabular}{lllrccccccc}
\toprule
Dataset & Model & Size & Params & Window F1 & Stream F1 & Agreement & Latent error & Step, CPU & Step, GPU & Window, GPU \\
\midrule
FI-2010 & Continual Transformer & mini & 971 & 0.564 & 0.564 & 1.000 & 3.6e-07 & 0.29 (0.00) & 0.69 (0.00) & 0.70 (0.00) \\
 &  & small & 2,963 & 0.603 & 0.603 & 1.000 & 3.4e-07 & 0.27 (0.00) & 0.64 (0.00) & 0.68 (0.00) \\
 &  & tiny & 18,563 & 0.654 & 0.654 & 1.000 & 3.2e-07 & 1.23 (0.00) & 1.88 (0.00) & 1.22 (0.00) \\
 &  & base & 69,891 & 0.684 & 0.684 & 1.000 & 3.4e-07 & 1.29 (0.00) & 1.80 (0.00) & 1.14 (0.00) \\
 &  & large & 1,065,987 & 0.660 & 0.660 & 1.000 & 3.8e-07 & 3.13 (1.00) & 1.67 (0.00) & 1.07 (0.00) \\
 &  & xlarge & 4,229,123 & 0.671 & 0.671 & 1.000 & 3.4e-07 & 8.06 (1.00) & 1.95 (0.00) & 1.28 (0.00) \\
 & Mamba-2 (recurrent) & mini & 1,111 & 0.618 & 0.615 & 0.984 & 2.9e-02 & 0.30 (0.00) & 0.62 (0.00) & 0.97 (0.00) \\
 &  & small & 2,895 & 0.646 & 0.646 & 0.995 & 9.6e-03 & 0.29 (0.00) & 0.59 (0.00) & 0.94 (0.00) \\
 &  & tiny & 33,011 & 0.682 & 0.677 & 0.964 & 1.0e-01 & 1.00 (0.00) & 1.98 (0.00) & 3.59 (0.19) \\
 &  & base & 245,827 & 0.655 & 0.648 & 0.951 & 1.4e-01 & 2.11 (0.00) & 3.62 (1.00) & 6.22 (0.52) \\
 &  & large & 6,924,163 & 0.675 & 0.672 & 0.972 & 1.0e-01 & 5.61 (1.00) & 6.80 (1.00) & 11.88 (0.75) \\
 &  & xlarge & 33,102,275 & 0.671 & 0.669 & 0.963 & 1.2e-01 & 22.73 (1.00) & 10.78 (1.00) & 20.87 (0.86) \\
\midrule
Synthetic-1 & Continual Transformer & mini & 763 & 0.463 & 0.463 & 1.000 & 1.1e-07 & 0.29 (0.00) & 0.63 (0.00) & 0.66 (0.00) \\
 &  & small & 2,547 & 0.462 & 0.462 & 1.000 & 1.6e-07 & 0.29 (0.00) & 0.63 (0.00) & 0.66 (0.00) \\
 &  & tiny & 17,731 & 0.919 & 0.919 & 1.000 & 2.8e-07 & 1.41 (0.00) & 2.13 (0.00) & 1.33 (0.00) \\
 &  & base & 68,227 & 0.934 & 0.934 & 1.000 & 2.7e-07 & 1.32 (0.00) & 1.63 (0.00) & 1.02 (0.00) \\
 &  & large & 1,059,331 & 0.941 & 0.941 & 1.000 & 2.7e-07 & 3.13 (1.00) & 1.77 (0.00) & 1.14 (0.00) \\
 &  & xlarge & 4,215,811 & 0.938 & 0.938 & 1.000 & 3.2e-07 & 8.26 (1.00) & 2.27 (0.07) & 1.22 (0.00) \\
 & Mamba-2 (recurrent) & mini & 903 & 0.493 & 0.492 & 0.964 & 1.6e-01 & 0.33 (0.00) & 0.63 (0.00) & 0.97 (0.00) \\
 &  & small & 2,479 & 0.541 & 0.540 & 0.995 & 4.4e-03 & 0.32 (0.00) & 0.62 (0.00) & 0.96 (0.00) \\
 &  & tiny & 32,179 & 0.933 & 0.933 & 0.999 & 2.6e-03 & 0.97 (0.00) & 1.84 (0.00) & 3.17 (0.14) \\
 &  & base & 244,163 & 0.939 & 0.938 & 0.988 & 5.8e-02 & 2.50 (0.00) & 4.26 (1.00) & 8.32 (0.64) \\
 &  & large & 6,917,507 & 0.934 & 0.934 & 0.989 & 4.7e-02 & 6.00 (1.00) & 7.57 (1.00) & 13.41 (0.78) \\
 &  & xlarge & 33,088,963 & 0.931 & 0.931 & 0.989 & 6.3e-02 & 15.62 (1.00) & 8.55 (1.00) & 15.58 (0.81) \\
\midrule
Synthetic-2 & Continual Transformer & mini & 811 & 0.559 & 0.559 & 1.000 & 1.6e-07 & 0.26 (0.00) & 0.67 (0.00) & 0.68 (0.00) \\
 &  & small & 2,643 & 0.583 & 0.583 & 1.000 & 1.5e-07 & 0.30 (0.00) & 0.63 (0.00) & 0.65 (0.00) \\
 &  & tiny & 17,923 & 0.621 & 0.621 & 1.000 & 2.1e-07 & 1.25 (0.00) & 1.62 (0.00) & 1.03 (0.00) \\
 &  & base & 68,611 & 0.694 & 0.694 & 1.000 & 2.5e-07 & 1.44 (0.00) & 1.92 (0.00) & 1.21 (0.00) \\
 &  & large & 1,060,867 & 0.731 & 0.731 & 1.000 & 2.9e-07 & 3.35 (1.00) & 1.78 (0.00) & 1.15 (0.00) \\
 &  & xlarge & 4,218,883 & 0.750 & 0.750 & 1.000 & 3.8e-07 & 8.18 (1.00) & 2.10 (0.00) & 1.12 (0.00) \\
 & Mamba-2 (recurrent) & mini & 951 & 0.583 & 0.583 & 0.999 & 3.8e-03 & 0.28 (0.00) & 0.60 (0.00) & 0.94 (0.00) \\
 &  & small & 2,575 & 0.588 & 0.588 & 0.998 & 2.9e-03 & 0.28 (0.01) & 0.55 (0.00) & 0.88 (0.00) \\
 &  & tiny & 32,371 & 0.597 & 0.597 & 0.996 & 8.8e-03 & 0.90 (0.00) & 1.75 (0.00) & 3.19 (0.07) \\
 &  & base & 244,547 & 0.595 & 0.594 & 0.957 & 6.6e-02 & 2.04 (0.00) & 3.07 (1.00) & 5.70 (0.48) \\
 &  & large & 6,919,043 & 0.620 & 0.618 & 0.952 & 7.6e-02 & 8.14 (1.00) & 8.60 (1.00) & 15.94 (0.81) \\
 &  & xlarge & 33,092,035 & 0.588 & 0.587 & 0.983 & 2.0e-02 & 15.36 (1.00) & 8.98 (1.00) & 16.92 (0.83) \\
\bottomrule
\end{tabular}
}
\end{table}

\end{document}